%% file: sample.tex
\documentclass[twoside,11pt]{article}

\usepackage{blindtext}

\usepackage{dmlr2e}

\usepackage{mdframed}
\usepackage[acronym]{glossaries}
\usepackage{multirow}
\usepackage{enumitem}
\usepackage{pifont}
\newcommand{\xmark}{\ding{55}}%
\usepackage{textcomp}
\newcommand{\textcustomtilde}{\raisebox{0.5ex}{\texttildelow}}
\usepackage{subcaption}
\usepackage{wrapfig} 
\usepackage{booktabs}
\usepackage{dcolumn}
\usepackage{cleveref}

\Crefname{section}{Section}{Sections}
\crefname{section}{Section}{Sections}
\Crefname{figure}{Figure}{Figures}
\crefname{figure}{Figure}{Figures}
\Crefname{table}{Table}{Tables}
\crefname{table}{Table}{Tables}
\Crefname{appendix}{Appendix}{Appendices}
\crefname{appendix}{Appendix}{Appendices}

\PassOptionsToPackage{hyphens}{url}


\glsdisablehyper
\newacronym{ml}{ML}{Machine Learning}
\newacronym{nn}{NN}{Neural Network}
\newacronym{fpga}{FPGA}{Field Programmable Gate Array}
\newacronym{asic}{ASIC}{Application Specific Integrated Circuit}
\newacronym{vww}{VWW}{Visual Wake Words}
\newacronym{VWW}{VWW}{Visual Wake Words}
\newacronym{miap}{MIAP}{More Inclusive Annotations for People}
\newacronym{url}{URL}{Uniform Resource Locator}
\newacronym{coco}{COCO}{Common Objects in Context}
\newacronym{tfds}{TFDS}{TensorFlow Datasets}
\newacronym{bbox}{bbox}{Bounding Box}

\newcommand{\gcheckmark}{{\color{green}\checkmark}}
\newcommand{\rxmark}{{\color{red}\xmark}}

\newcommand{\rv}[1]{#1}

\newif\ifshowedits

\newcommand{\new}[1]{%
  \ifshowedits
    \textcolor{blue}{#1}
  \else
    #1
  \fi
}

\usepackage{lastpage}
\dmlrheading{26}{2026}{1-\pageref{LastPage}}{04/26}{04/26}{21-0000}{Banbury, Njor, Garavagno, Mazumder et al} 
\def\openreview{\url{https://openreview.net/forum?id=eb82tUIZRA}}

\ShortHeadings{Wake Vision}{Banbury, Njor, Garavagno, Mazumder et al.}
\firstpageno{1}

\begin{document}

\title{Wake Vision: A Tailored Dataset and Benchmark Suite for TinyML Computer Vision Applications}

\author{%
       \name Colby Banbury \email colbybanbury@gmail.com \\
       \addr Harvard University\\
       Cambridge, MA 02138, USA
       \AND
       \name Emil Njor\thanks{Now affiliated with The University of Southern Denmark, Vejle, 7100, Denmark} \email njor@mmmi.sdu.dk \\
       \addr Technical University of Denmark, Kongens Lyngby, 2800, Denmark\\
       Harvard University, Cambridge, MA 02138, USA
       \AND
       \name Andrea Mattia Garavagno \email AndreaMattia.Garavagno@edu.unige.it \\
       \addr Harvard University, Cambridge, MA 02138, USA \\
       Scuola Superiore Sant'Anna, Pisa, 56127, Italy \\ 
       University of Genoa, Genoa, 16126, Italy
       \AND
       \name Mark Mazumder \email markmazumder@g.harvard.edu \\
       \addr Harvard University\\
       Cambridge, MA 02138, USA
       \AND
       \name Matthew Stewart \email matthew\_stewart@g.harvard.edu \\
       \addr Harvard University\\
       Cambridge, MA 02138, USA
       \AND
       \name Pete Warden \email pete@petewarden.com \\
       \addr
       Moonshine AI
       \AND
       \name Manjunath Kudlur \\
       \addr
       Moonshine AI
       \AND
       \name Nat Jeffries \\
       \addr
       Moonshine AI
       \AND
       \name Xenofon Fafoutis \email xefa@dtu.dk \\
       \addr Technical University of Denmark\\
       Kongens Lyngby, 2800, Denmark
       \AND
       \name Vijay Janapa Reddi \email vj@eecs.harvard.edu \\
       \addr Harvard University\\
       Cambridge, MA 02138, USA
       }

\editor{Zach Xu}

\maketitle

\newpage

\begin{abstract}
\input{sec/0_abstract}
\end{abstract}


\input{sec/1_intro}
\input{sec/2_background}
\input{sec/3_wake_vision}

\input{sec/4_results}
\input{sec/5_ethics}

\input{sec/6_conclusion}


\acks{The Google TPU Research Cloud program partially supported this work via cloud computing credits. 
    The PhD stipend of Emil Njor was supported by the Innovation Fund Denmark DIREC project (9142-00001B).
    Emil Njor's external research stay was furthermore supported by Fulbright Denmark, Stibo Fonden, Thomas. B. Thriges Fond, Otto Mønsteds Fond and Kaj og Hermilla Ostenfelds Fond.
    This work was also supported by the National Science Foundation (NSF) and the Semiconductor Research Corporation (SRC).}

\vskip 0.2in
\bibliography{bib}

\input{sec/appendix}
\input{sec/datasheet}

\end{document}

%% file: sec/0_abstract.tex
Tiny machine learning (TinyML) co-locates models with sensors on microcontrollers, where small models (which are disproportionately sensitive to label noise) and bespoke binary tasks (which lack standard benchmarks) make general-purpose dataset practices a poor fit. 
\acrfull{vww}, the prior standard TinyML person detection benchmark, contains roughly 123K images and has an estimated label error rate of 7.8\%, which limits its usefulness for production-grade systems. Manual labeling, however, is prohibitively expensive for the scale and diversity of TinyML use cases.

We address this gap with the Wake Vision pipeline, an automated method for generating and curating large-scale binary classification datasets for TinyML. 
We use \emph{data-centric TinyML} for the dataset construction, curation, and lifecycle methods that produce the large, well-curated datasets these systems require. 
The pipeline combines label fusion across image-level and bounding-box sources, confidence-, area-, and depiction-aware filtering, label correction on the evaluation splits, and automatic generation of fine-grained benchmark subsets. Applying it to person detection, we release Wake Vision, a dataset of almost 6M images (close to 100$\times$ more person images than VWW) with a manually relabeled validation and test set at a 2.2\% label error rate.

Models trained on Wake Vision improve test accuracy by up to 6.6\% over VWW across MobileNetV2, MCUNet, MicroNets, and ColabNAS architectures, and match or exceed VWW-trained models on 13 of 16 fine-grained subsets covering perceived gender, perceived age, distance, lighting, and depictions. The advantage holds under distribution shift on three out-of-distribution datasets covering driving and overhead-surveillance imagery. 
We additionally uncover two TinyML-specific insights: small models are more sensitive to label errors than large models, and two-stage training, which pretrains on the noisier large set and fine-tunes on the cleaner small set, is a viable strategy even for tiny, low-capacity models. 
Beyond person detection, the Wake Vision pipeline applies to the 9.6K trainable classes of Open Images v7; on bird detection it produces a dataset 27$\times$ larger than a VWW-style baseline with label error reduced from 6.6\% to 0.6\%. All artifacts are released under CC-BY 4.0 through TensorFlow Datasets and Hugging Face. To continue improving the dataset over time, we partner with the Edge AI Foundation to host community competitions; the first round contributed a label-correction technique that reduced the Wake Vision (Large) label error rate from 15.2\% to 9.8\%, at a cost orders of magnitude below the \$600,000 implied by manual relabeling at this scale.


%% file: sec/1_intro.tex
\section{Introduction}\label{sec:introduction}

Tiny machine learning (TinyML)~\citep{banbury2020benchmarking, warden2019tinyml, banbury2021mlperf} co-locates models with sensors on microcontrollers (MCUs), typically with a memory budget of hundreds of kilobytes, nearly four orders of magnitude less than common smartphone models~\citep{banbury2021micronets, david2021tensorflow}.
These constraints rule out conventional benchmarks. ImageNet's thousand-class output is infeasible on MCUs~\citep{deng2009imagenet, chowdhery2019visual}, and training data must instead come from datasets purpose-built for the binary or low-class-count tasks that fit MCU budgets.
Existing TinyML datasets such as \gls{vww}~\citep{chowdhery2019visual} and Google Speech Commands~\citep{warden2018speech} target exactly these simpler applications, supporting binary classification and tens of classes, respectively, but their limited scale and label error rates make them unsuitable for training and evaluating production-grade TinyML models.

The gap is methodological as well as empirical. Building TinyML datasets at the scale required for modern training, and across the diversity of tasks TinyML is being deployed for, requires automated pipelines for creating, curating, and evaluating datasets, not just larger downloads. We use \emph{data-centric TinyML} for dataset construction, curation, and lifecycle methods that target two TinyML-specific constraints: (i) deployment models small enough where label noise materially erodes accuracy 
and (ii) bespoke binary or low-class-count tasks for which standard benchmarks do not exist, requiring datasets to be generated per-task from broader sources rather than reused, a property the Wake Vision pipeline addresses by targeting any of the 9.6K trainable classes of Open Images v7 (\Cref{sec:other_binary_datasets}).

\begin{table*}[t!]
    \centering
    \caption{Wake Vision (\textbf{bold}) compared against the standard TinyML person detection benchmark and two general image classification datasets sometimes used as person-detection proxies~\citep{chowdhery2019visual,krizhevsky2009learning,pascal-voc-2012}. Wake Vision (Large) and Wake Vision (Quality) share the same validation and test sets.}
    \label{tab:person-detection-datasets}
    \small
    \setlength{\tabcolsep}{2pt}
    \begin{tabular}{lD{.}{.}{0}D{.}{.}{0}D{.}{.}{0}D{.}{.}{0}cc}
        \toprule
        & \multicolumn{1}{c}{Total} & \multicolumn{3}{c}{\# of Person Images} & Fine-Grained & Suitable for\\
        Dataset & \multicolumn{1}{c}{Images} & \multicolumn{1}{c}{Train} & \multicolumn{1}{c}{Validation} & \multicolumn{1}{c}{Test} & Filtering & TinyML\\
        \midrule
        \textbf{Wake Vision (Quality)} & \textbf{1,322,574} & \textbf{624,115} & \textbf{9,291} & \textbf{27,881} & \textbf{\gcheckmark} & \textbf{\gcheckmark} \\
        \textbf{Wake Vision (Large)} & \textbf{5,760,428} & \textbf{2,880,214} & - & - & \textbf{\rxmark} & \textbf{\gcheckmark} \\\midrule
        Visual Wake Words & 123,287 & 36,000 & 3,926 & 19,107 & \textbf{\rxmark} & \textbf{\gcheckmark} \\ 
        CIFAR-100 & 60,000 & 2,500 & - & 500 & \textbf{\rxmark} & \textbf{\rxmark} \\
        PASCAL VOC 2012 & 11,530 & 1,994 & 2,093 & - & \textbf{\rxmark} & \textbf{\rxmark} \\
        \bottomrule
    \end{tabular}
\end{table*}

Wake Vision is the successor to \acrfull{vww}. It preserves the binary person-detection task and evaluation conventions VWW established for the TinyML community, and addresses VWW's limitations in scale, label quality, and licensing through the Wake Vision pipeline, an automated method for generating and curating large-scale binary classification datasets for TinyML. The pipeline fuses multiple label sources, applies confidence-, area-, and depiction-aware filtering, supports manual and automated label correction on the evaluation splits, and generates fine-grained benchmark subsets (\Cref{fig:wake-vision-overview} in \Cref{sec:dataset_development}).
Automation is a necessity rather than a convenience. Our manual relabeling of approximately 70K validation and test images cost \$7,000 USD, which extrapolates to roughly \$600,000 USD for a 6M-image training set, and human labelers introduce inconsistency at scale.
The pipeline derives datasets from Open Images v7~\citep{OpenImages, OpenImages2}, which lets the resulting datasets be released under a permissive CC-BY 4.0 license.

\begin{figure*}[!htbp]
    \centering
    \subcaptionbox{Older Person}{\includegraphics[height=2cm]{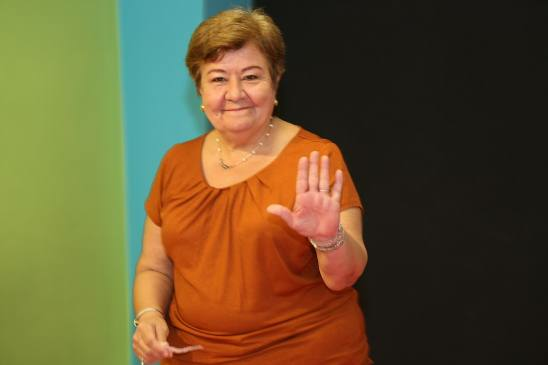}}
    \hfill
    \subcaptionbox{Near Person}{\includegraphics[height=2cm]{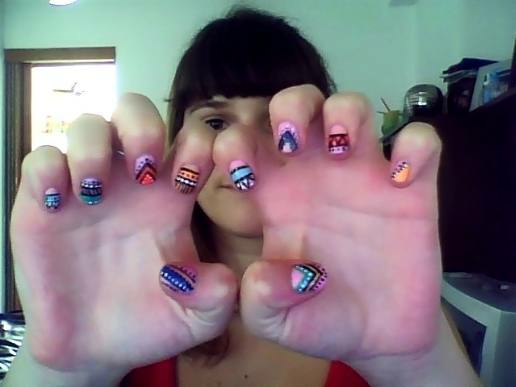}}
    \hfill
    \subcaptionbox{Bright Image}{\includegraphics[height=2cm]{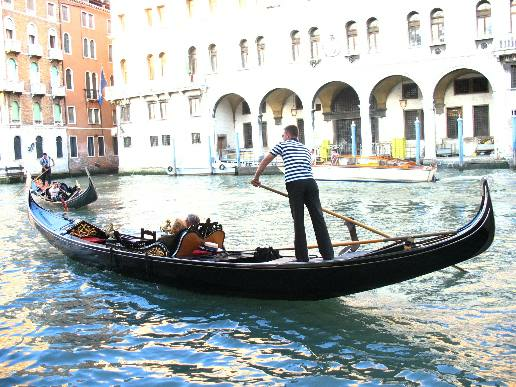}}
    \subcaptionbox{Female Person}{\includegraphics[height=2cm]{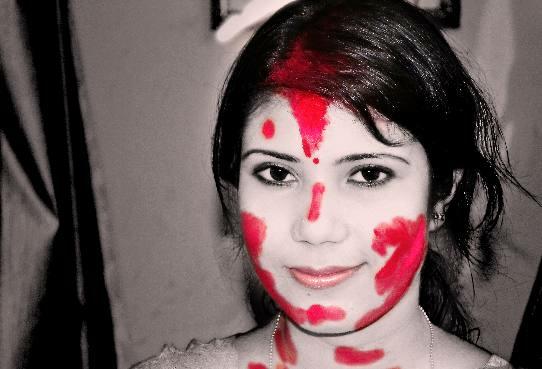}}
    \hfill
    \subcaptionbox{Depicted Person}{\includegraphics[height=2cm]{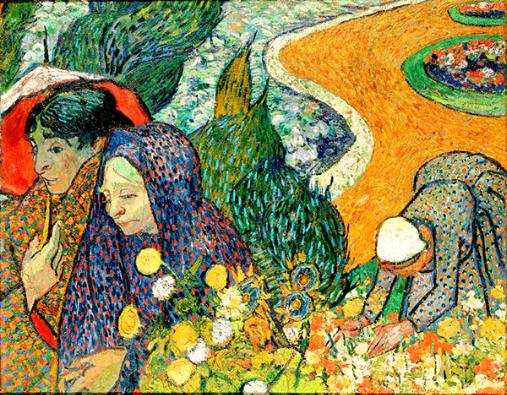}}
    \hfill
    \caption{One sample from each of the five fine-grained benchmark dimensions in Wake Vision: perceived age, distance, lighting, perceived gender, and depictions. Full subset compositions appear in \Cref{fig:extensive_fine_grained_benchmark_examples}.}
    \label{fig:wake_vision_samples}
\end{figure*}

We demonstrate the pipeline by creating Wake Vision, a binary person detection dataset. Person detection is the canonical vision use case for TinyML~\citep{banbury2021mlperf}, supporting applications from occupancy detection~\citep{piechocki2022efficient} and smart HVAC\slash lighting~\citep{zacharia2022intelligent} to an always-on `wake model' that gates power-intensive sensors and larger models~\citep{heitz2008cascaded}.
As shown in Table~\ref{tab:person-detection-datasets}, Wake Vision provides almost 6M images, close to 100$\times$ more person images than \acrfull{vww}~\citep{chowdhery2019visual}, and improves test accuracy by up to 6.6\%.
Alongside the dataset, we develop a suite of five fine-grained benchmarks (perceived gender, perceived age, distance, lighting, and depictions; \Cref{fig:wake_vision_samples}, \Cref{sec:benchmark_suite}), on which Wake Vision-trained models match or exceed \gls{vww}-trained models on 13 of 16 subsets, exposing failure modes that an aggregate accuracy metric obscures.

Wake Vision exposes two TinyML-specific findings (\Cref{sec:training-set-eval},~\Cref{sec:quality_vs_size}). First, smaller models lose more accuracy from training-set label noise than larger models do. Under our label-noise injection protocol, the smallest model loses 1.3\% accuracy when the training error rate rises from 6.8\% to 15\%, versus 0.5\% for the largest. Second, two-stage training that pretrains on Wake Vision (Large) and fine-tunes on Wake Vision (Quality) reaches 85.72\% test accuracy, versus 84.89\% for Quality alone and 80.80\% for Large alone, combining the scale of one set and the quality of the other.

Our work makes four contributions:

\begin{itemize}[nosep]

\item \textbf{The Wake Vision pipeline}, an automated method for building bespoke large-scale binary classification datasets for TinyML, combining multi-source label fusion, confidence-, area-, and depiction-aware filtering, label correction, and fine-grained benchmark synthesis (Sections~\ref{sec:labeling}--\ref{sec:benchmark_suite}). Beyond person detection, we apply the pipeline to two additional Open Images categories chosen because they also exist in MS-COCO, allowing direct comparison to the \gls{vww}-derived methodology: it produces a car detection dataset 12$\times$ larger than the VWW-style baseline (label error 6.4\% $\to$ 4.8\%) and a bird detection dataset 27$\times$ larger (6.6\% $\to$ 0.6\%, both estimated on 500-image samples), letting TinyML practitioners generate task-specific datasets without further large-scale labeling effort (\Cref{sec:other_binary_datasets},~\Cref{sec:birds_cars_results}).

\item \textbf{Wake Vision and a community-driven improvement loop.} A person detection dataset of almost 6M images (close to 100$\times$ more person images than \gls{vww}), with a manually relabeled validation and test set at a 2.2\% label error rate (versus 7.8\% in VWW), released under CC-BY 4.0 through TensorFlow Datasets and Hugging Face. We treat the dataset as a maintained artifact through recurring competitions hosted by the Edge AI Foundation; the first round contributed a label-correction technique that reduced the Wake Vision (Large) label error rate from 15.2\% to 9.8\%, a single-round improvement to a property that conventionally requires manual relabeling at a cost we estimate at \$600,000 for a 6M-image set (Sections~\ref{sec:label_correction},~\ref{sec:wake_vision_challenge}).

\item \textbf{A fine-grained benchmark suite} that evaluates TinyML model robustness across five dimensions: perceived gender, perceived age, distance, lighting, and depictions. Wake Vision-trained models match or exceed \gls{vww}-trained models on 13 of 16 fine-grained subsets (strict wins on 11, ties on 2 lighting subsets), with the largest gains on depicted persons, older individuals, and dark lighting (\Cref{sec:benchmark_suite},~\Cref{sec:fine_grained_datasets.benchmark_results}).

\item \textbf{Empirical findings for data-centric TinyML}, including that two-stage training (Wake Vision Large pretraining followed by Wake Vision Quality fine-tuning) achieves the best accuracy across MobileNetV2, MCUNet, MicroNets, and ColabNAS architectures, that small models are disproportionately sensitive to label noise, and that the accuracy advantage of Wake Vision-trained models persists under distribution shift on driving and overhead-surveillance imagery (\Cref{sec:training-set-eval},~\Cref{sec:quality_vs_size},~\Cref{sec:ood}).

\end{itemize}

Our dataset, benchmarks, code, and models are at \url{https://wakevision.ai}.

%% file: sec/2_background.tex
\section{Background and Related Work}\label{sec:related_work}

We organize related work into three categories:

\paragraph{Generating Binary Classification Datasets for TinyML}
\label{sec:datasets_generation}
Automatically generated datasets can significantly lower costs associated with deploying TinyML systems. 
Prior to our work, the primary method for generating binary classification datasets for TinyML applications was through the code used to create \acrfull{vww}~\citep{chowdhery2019visual}, which can be applied to other MS-COCO~\citep{lin2014microsoft} classes.
While we offer the same capability for Open Images, the Wake Vision pipeline extends beyond dataset generation, providing customizable curation properties that include label correction, fine-grained benchmark suite generation, and community-driven enhancements through competitions, addressing the entire dataset lifecycle.

\paragraph{Person Detection in TinyML Applications}
\label{sec:pd}
Person detection has emerged as the canonical computer vision use case for TinyML systems~\citep{banbury2020benchmarking, banbury2021mlperf}. 
In the context of resource-constrained computing, TinyML systems demand tasks that carefully balance computational feasibility with real-world utility. 
Person detection represents an optimal compromise. It is sufficiently lightweight to be implemented on embedded microcontrollers while maintaining enough discriminative power for practical applications.
In particular, person detection serves as a critical activation mechanism in resource-constrained systems. 
These lightweight models support always-on deployment on low-powered MCUs to selectively ``wake'' power-intensive sensors, processors, and larger, more capable ML models upon detecting human presence~\citep{heitz2008cascaded}. 
This architectural pattern enables substantial energy savings while preserving privacy through on-device processing, and supports diverse applications ranging from occupancy detection~\citep{piechocki2022efficient}, smart HVAC systems~\citep{zacharia2022intelligent}, or poacher detection~\citep{doull2021evaluation}.

\paragraph{Existing Person Detection Datasets}
\label{sec:datasets}
In the TinyML computer vision domain, the \acrfull{vww} dataset~\citep{chowdhery2019visual, david2021tensorflow, lin2020mcunet, banbury2021micronets, banbury2021mlperf} has established itself as the de facto standard. 
Before Wake Vision, it was the only open-source dataset specifically designed for person detection with direct commercial applicability. 
However, \gls{vww} faces significant limitations: its small size and indirect access requirements (requiring regeneration from MS-COCO~\citep{lin2014microsoft}) restrict its accessibility and utility.

Although other datasets contain person labels within general image classification collections, such as Cifar-100~\citep{krizhevsky2009learning} and PASCAL Visual Object Classes~\citep{pascal-voc-2012}, these present significant drawbacks for TinyML applications. 
Specifically, their inadequate representation of the open ``no-person'' class can lead to poor perceived performance of TinyML models~\citep{chowdhery2019visual}. 
Our Wake Vision dataset addresses these limitations by providing nearly two orders of magnitude more images than any existing public, permissively licensed dataset (\Cref{tab:person-detection-datasets}). 
Furthermore, it distinguishes itself as the only person detection dataset offering a fine-grained benchmark suite and official distribution through popular services like \gls{tfds} and Hugging Face Datasets.

%% file: sec/3_wake_vision.tex
\section{Person Detection Dataset and Benchmark Generation}\label{sec:dataset_development}
\begin{figure}[t]
    \centering
    \includegraphics[scale=1]{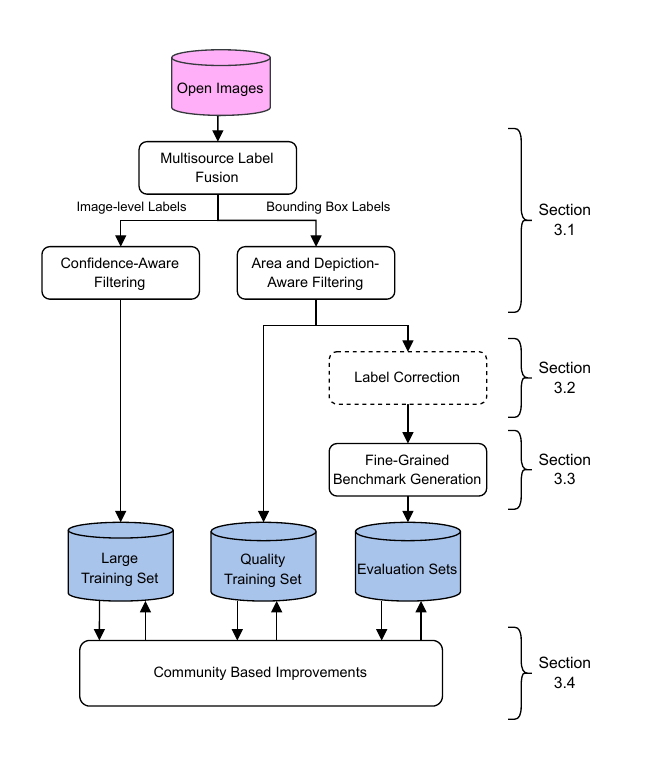}
    \caption{The Wake Vision dataset generation pipeline. Open Images image-level and bounding-box annotations are fused, filtered for confidence, person bounding-box area, and depiction status, optionally relabeled (manually or automatically), and split into training, validation, and test sets together with fine-grained benchmark subsets. Stages are detailed in \Cref{sec:labeling,sec:label_correction,sec:benchmark_suite}; dotted lines indicate optional steps.}
    \label{fig:wake-vision-overview}
\end{figure}

We apply the Wake Vision pipeline shown in \cref{fig:wake-vision-overview} to person detection, producing a \emph{production-grade} dataset two orders of magnitude larger than prior TinyML person detection datasets (\cref{tab:wv_size}). We use \emph{production-grade} to mean a dataset whose scale, manually relabeled evaluation splits, permissive license, and distribution through standard ML frameworks make it suitable for training and evaluating models that will be deployed in real systems, rather than only used as a research benchmark. The dataset's size, quality, and detailed metadata open avenues of TinyML research that smaller benchmarks cannot, and \cref{fig:wake_vision_samples} illustrates example images.

More specifically, the pipeline implements six data-centric operations (\Cref{fig:wake-vision-overview}): (1) \emph{multi-source label fusion} that reconciles image-level and bounding-box annotations from Open Images into a single binary task (\Cref{sec:labeling}); (2) \emph{confidence-, area-, and depiction-aware filtering} that removes uncertain, distant-subject, and depicted-only samples to match deployment conditions (\Cref{sec:labeling}); (3) \emph{automated label correction} via Confident Learning on the evaluation splits as a low-cost first pass (Appendix~\ref{app:auto-label-correction}); (4) \emph{manual relabeling} of the validation and test sets through a crowd-sourced platform, prioritized to where label quality has the largest effect on reported model performance (\Cref{sec:label_correction}); (5) \emph{fine-grained benchmark synthesis} that uses the per-image metadata produced by steps (1)--(4) to derive evaluation subsets along axes that an aggregate accuracy metric obscures (\Cref{sec:benchmark_suite}); and (6) \emph{community-driven continuous improvement} through recurrent competitions whose winning contributions are integrated into subsequent dataset releases (\Cref{sec:wake_vision_challenge}). Sections~\ref{sec:labeling}--\ref{sec:wake_vision_challenge} apply these operations to person detection, and \Cref{sec:other_binary_datasets} demonstrates that the same pipeline targets the 9.6K trainable classes of Open Images v7, with a comparison to \acrfull{vww}.

\subsection{Label Generation}\label{sec:labeling}
A large person detection dataset is indispensable for TinyML research. 
However, manually labeling millions of images would be prohibitively expensive for a nascent field like TinyML. 
Therefore, we use the Wake Vision pipeline to bootstrap Wake Vision using existing large-scale data efforts.

The base label in Wake Vision is a binary person/non-person label.
The first step in the Wake Vision pipeline is to derive these labels from the Open Images dataset~\citep{OpenImages, OpenImages2}, which contains both image-level and bounding box labels of 9.6K trainable classes and 600 objects, respectively.
Image-level labels are unlocalized, describing objects present in an image, whereas bounding box labels localize an object by four coordinates. 
The bounding box label classes are hierarchically structured, allowing one class to be a subcategory or a part of another class. 
For example a ``Woman'' is a subcategory of a ``Person,'' and a ``Human Hand'' is a part of ``Person.'' 
A comprehensive list of Open Images labels related to Wake Vision can be found in Appendix~\ref{sec:person_label_classes}.

Users can adapt the Wake Vision pipeline to meet the needs of specific use cases by utilizing the numerous configuration options available.
For example, a configuration option exists to change whether artistic depictions of humans are labeled as non-persons or excluded from the dataset.
We refer to Appendix~\ref{app:label_generation_details} for details about these configuration options.

We observe that bounding box labels are generally more accurate and provide more information for data filtering while being less numerous. The Open Images training set has \textcustomtilde{}9 million images with image-level person labels, but only \textcustomtilde{}1.7 million images with a person bounding box.
Consequently, this presents a trade-off between more data (image-level labels) or higher-quality labels (bounding box labels).
In response, we provide two Wake Vision training sets: Wake Vision (Large), labeled via Open Images image-level labels, and Wake Vision (Quality), a smaller set labeled via Open Images bounding boxes (Table~\ref{tab:person-detection-datasets}).
Since the Open Images validation and test sets are fully labeled with higher-quality bounding box labels, we derive the Wake Vision validation and test sets from the bounding box labels in these Open Images splits.

\begin{table*}
    \centering
    \caption{Number of images in the Wake Vision dataset}
    \label{tab:wv_size}
    \begin{tabular}{lccc}
        \toprule
         & Person Images & Non-Person Images & Excluded \\
        \midrule
        Wake Vision (Large) Training Dataset & 2,880,214 & 2,880,214 & 2,176,551 \\
        Wake Vision (Quality) Training Dataset & 624,115 & 624,115 & 6,688,749 \\
        Validation Dataset & 9,291 & 9,291 & 17,824 \\
        Test Dataset & 27,881 & 27,881 & 53,543 \\
        \bottomrule
    \end{tabular}
\end{table*}

\paragraph{Wake Vision (Large) Training Set}\label{sec:image_level_labeling}
Image-level labels in Open Images include a confidence property, which represents the certainty that a label is correct.
This confidence ranges from 0 to 10 for machine-generated labels and is strictly either 0 or 10 for human-verified labels.
See \cref{sec:label_correction} and Appendix~\ref{app:label_generation_details} for more information.
The Wake Vision pipeline can be configured to remove low-confidence machine-labeled person images, which we do for Wake Vision by only including labels with a minimum confidence of 7.

\begin{figure}[t]
    \centering
    \includegraphics[width=1.0\linewidth]{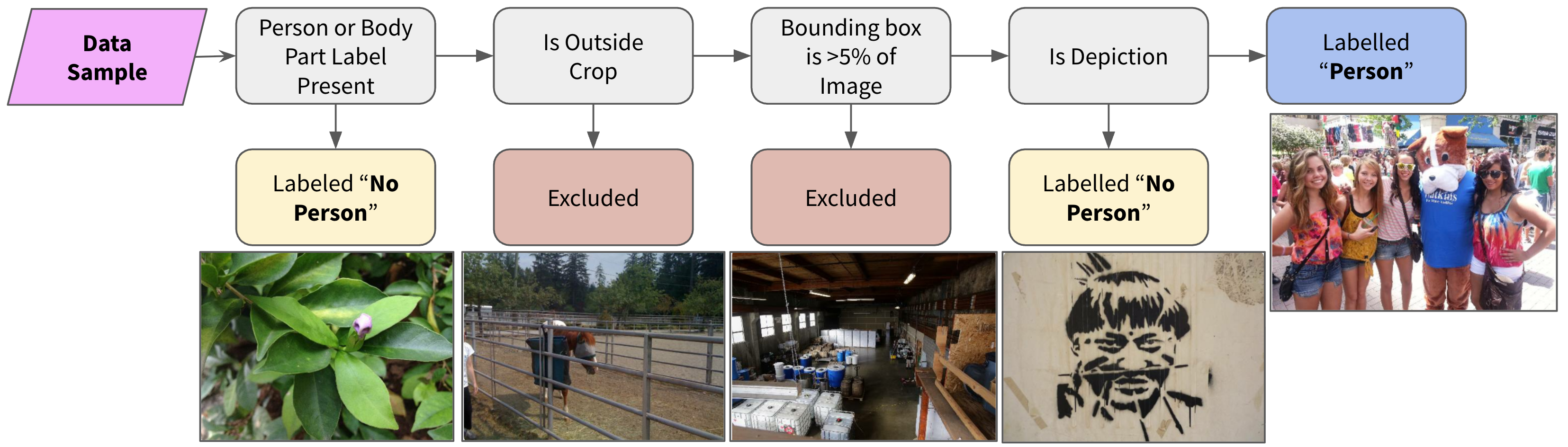}
    \caption{A flowchart of the bounding-box filtering process of an image for person detection.}
    \label{fig:filter-flowchart}
\end{figure}

\begin{figure}[t]
    \centering    
    \subcaptionbox{An example of a person far away\label{fig:far_away_person}}{\includegraphics[width=0.4\linewidth]{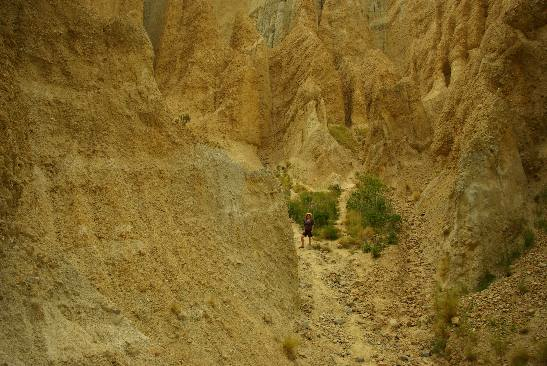}}
    \hspace{1cm}
    \subcaptionbox{An example of a depicted person\label{fig:depiction}}{\includegraphics[width=0.4\linewidth]{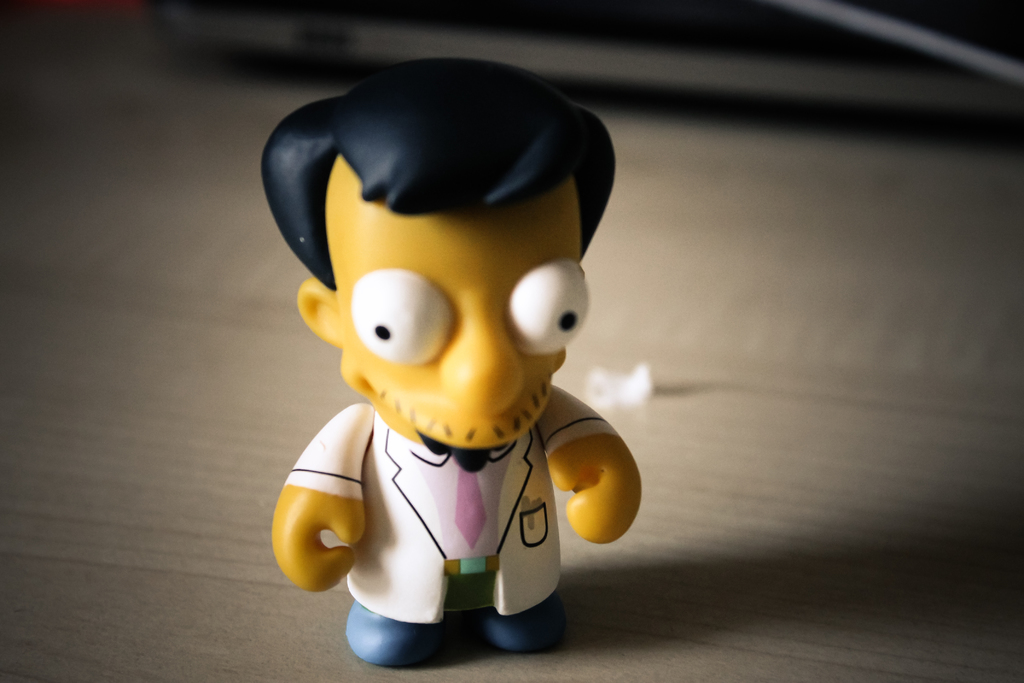}}
    \caption{Examples of challenging outlier images.}
\end{figure}

\paragraph{Wake Vision (Quality) Training Set}\label{sec:bbox_labeling}
Bounding box labels in Open Images, in contrast to image-level labels, are all verified and localized by humans, minimizing false positive labels.
Bounding box labels can be used to calculate an approximation of the area of an image occupied by a person.
We use area as a proxy for the distance of a person to the camera, and by default, exclude far-away persons, i.e., those that take up less than 5\% of the image (see ~\cref{fig:far_away_person} for an example).
Finally, bounding box attributes include an artistic depiction flag (an example is provided in ~\cref{fig:depiction}).
The Wake Vision pipeline excludes depictions from class members by default; that is, a depiction of a person is not considered a person.
See \cref{fig:filter-flowchart} for an illustration of the bounding box filtering process for person detection.

\subsection{Label Correction}
\label{sec:label_correction}
Label errors are a challenge in computer vision that limit progress and obscure true performance~\citep{northcutt2021pervasive, beyer2020we}.
This is especially prevalent in large datasets that use complex labeling pipelines to scale while managing costs.
Recognizing the importance of label quality, we take measures to estimate and improve the accuracy of Wake Vision's labels.

As any labels generated by the Wake Vision pipeline are derived from Open Images v7, errors in these labels are inherited by our datasets.
All image-level labels in the Open Images dataset originate from machine-generated candidate labels.
Many of these labels are later verified by human annotators, and a subset of them are given bounding boxes and additional annotations.

While the machine-generated label phase aims to identify objects efficiently, any instances missed during this initial step are unlikely to be captured in downstream phases. 
Consequently, the Open Images dataset may contain numerous false negative labels, referring to images where an object is present but lacks the corresponding label. 
Therefore, additional measures are necessary to identify and correct such labeling omissions. Given the size of datasets such as Wake Vision, manually correcting label errors across the entire dataset becomes an arduous undertaking.

For our Wake Vision dataset, we therefore prioritize labeling the validation and test sets so that the dataset can be used to accurately evaluate a model's performance.
For this task, we use the Scale Rapid tool from Scale AI (\url{https://scale.com/}) to relabel the Wake Vision validation and test sets.
Details about the relabeling process, including instructions provided and the total cost, can be found in Appendix~\ref{app:manual-label-correction}. 
With the ground truth established in the validation and test sets, Wake Vision becomes an ideal foundation for research on automated data cleaning techniques~\citep{northcutt2021confident}. 
We perform an initial exploration of these methods in Appendix~\ref{app:auto-label-correction}.

We report our estimated label error rate after label corrections versus \gls{vww} in Table~\ref{tab:wv-vww-test-errors}.
These estimates are all based on a random subset of 500 samples from each dataset, manually evaluated by the authors.
The Wake Vision validation and test set estimated error rates are considerably lower than that of the \gls{vww} dataset top-level error rate.

\begin{table}[t]
    \centering
    \caption{Label error rate of \gls{vww} and Wake Vision \rv{after label correction}. The Wake Vision Train (Quality) and Train (Large) sets have different error rates due to the different label sources. The Wake Vision Validation and Test sets have a lower error rate due to manual relabeling.}
    \label{tab:wv-vww-test-errors}
    \begin{tabular}{llc}
        \toprule
        Dataset & Wake Vision Set & Label Error\\
        \midrule
        Visual Wake Words & Train, Val, \& Test & 7.8\%\\
        \midrule
        Wake Vision & Train (Large) &  15.2\%\\
        Wake Vision & Train (Quality) &  6.8\%\\
        Wake Vision & Val \& Test & 2.2\% \\
        \bottomrule
    \end{tabular}
\end{table}

\begin{figure*}[t]
    \begin{minipage}[c]{0.15\linewidth}
        \centering
        \textbf{Gender}
    \end{minipage}
    \begin{subfigure}[c]{0.26\linewidth}
        \centering
        \includegraphics[height=1.8cm]{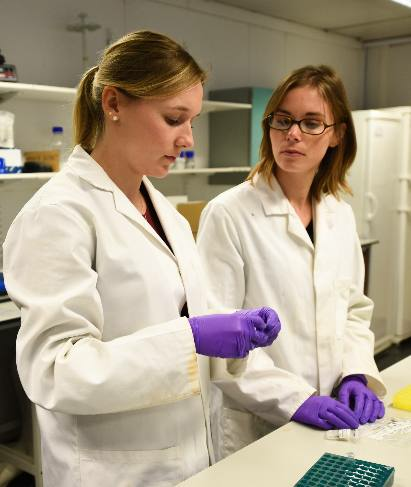}
        \caption*{Female}
    \end{subfigure}
    \begin{subfigure}[c]{0.26\linewidth}
        \centering
        \includegraphics[height=1.8cm]{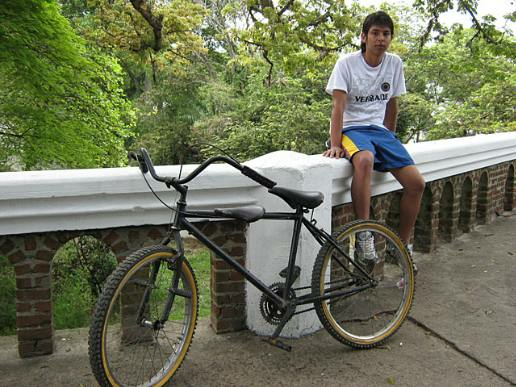}
        \caption*{Male}
    \end{subfigure}
    \begin{subfigure}[c]{0.26\linewidth}
        \centering
        \includegraphics[height=1.8cm]{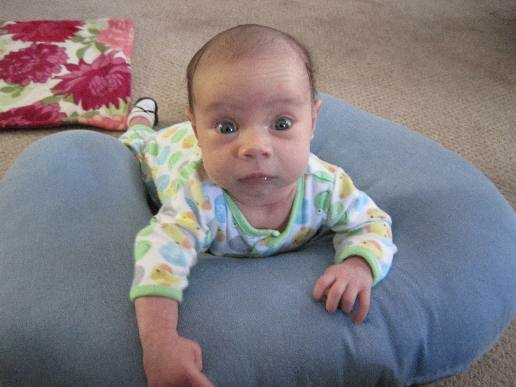}
        \caption*{Gender Unknown}
    \end{subfigure}
    \\
    \begin{minipage}[c]{0.15\linewidth}
        \centering
        \textbf{Age}
    \end{minipage}
    \begin{subfigure}[c]{0.15\linewidth}
        \centering
        \includegraphics[height=1.8cm]{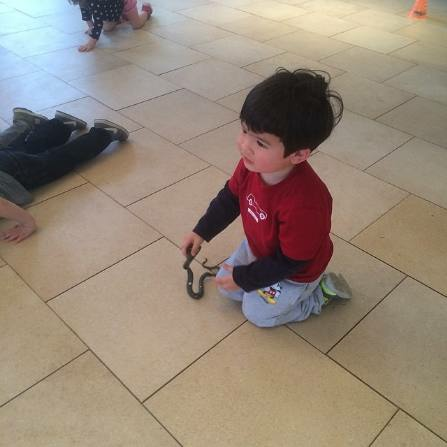}
        \caption*{Young}
    \end{subfigure}
    \begin{subfigure}[c]{0.23\linewidth}
        \centering
        \includegraphics[height=1.8cm]{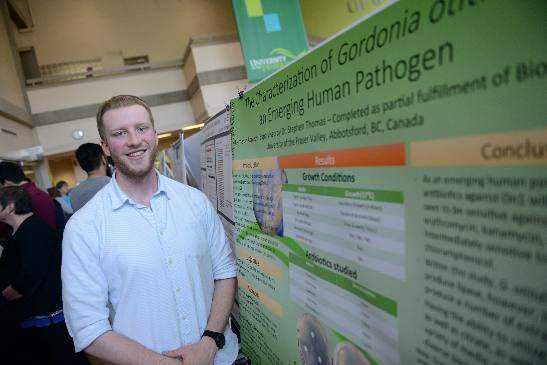}
        \caption*{Middle}
    \end{subfigure}
    \begin{subfigure}[c]{0.2\linewidth}
        \centering
        \includegraphics[height=1.8cm]{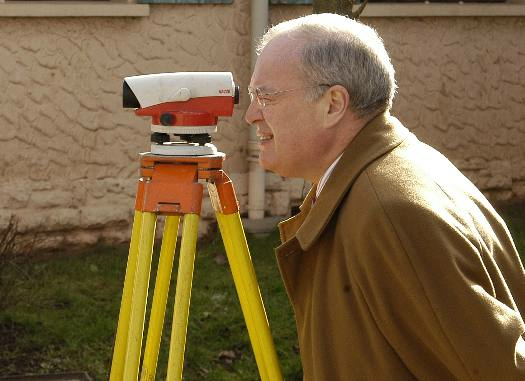}
        \caption*{Older}
    \end{subfigure}
    \begin{subfigure}[c]{0.23\linewidth}
        \centering
        \includegraphics[height=1.8cm]{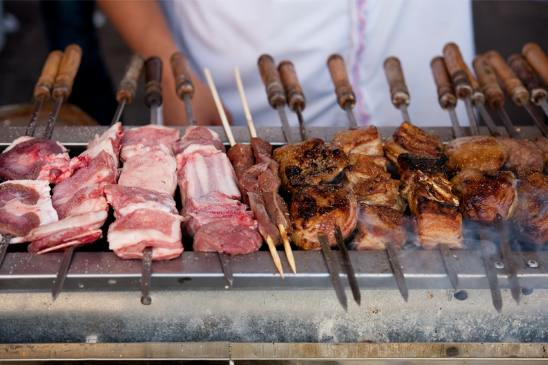}
        \caption*{Age Unknown}
    \end{subfigure}
    \\
    \begin{minipage}[c]{0.15\linewidth}
        \centering
        \textbf{Distance}
    \end{minipage}
    \begin{subfigure}[c]{0.26\linewidth}
        \centering
        \includegraphics[height=1.8cm]{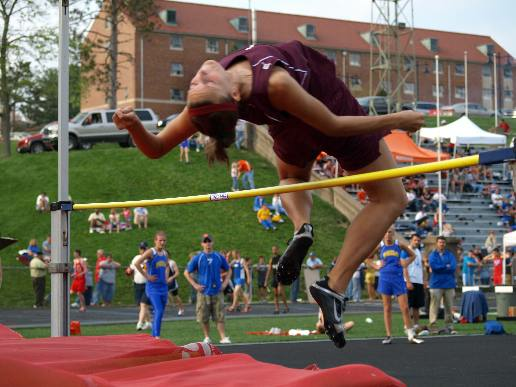}
        \caption*{Near}
    \end{subfigure}
    \begin{subfigure}[c]{0.26\linewidth}
        \centering
        \includegraphics[height=1.8cm]{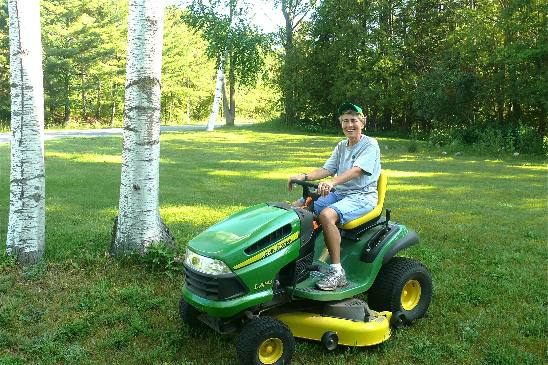}
        \caption*{Medium}
    \end{subfigure}
    \begin{subfigure}[c]{0.26\linewidth}
        \centering
        \includegraphics[height=1.8cm]{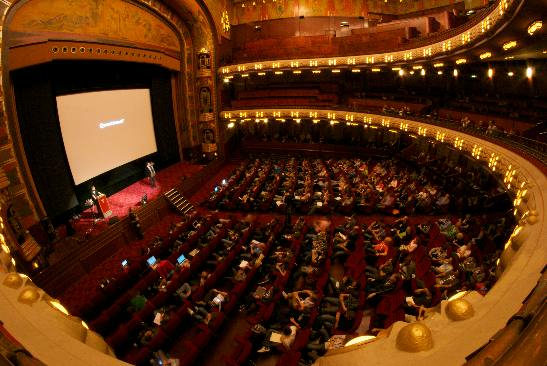}
        \caption*{Far}
    \end{subfigure}
    \\
    \begin{minipage}[c]{0.15\linewidth}
        \centering
        \textbf{Lighting}
    \end{minipage}
    \begin{subfigure}[c]{0.26\linewidth}
        \centering
        \includegraphics[height=1.8cm]{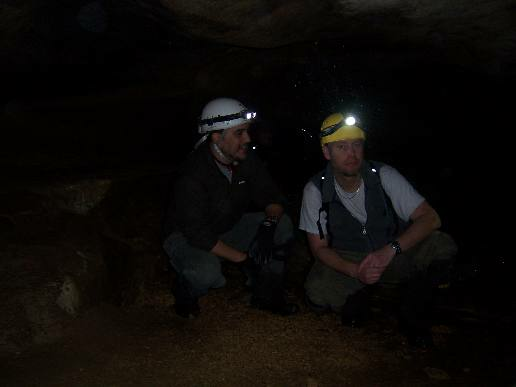}
        \caption*{Dark}
    \end{subfigure}
    \begin{subfigure}[c]{0.26\linewidth}
        \centering
        \includegraphics[height=1.8cm]{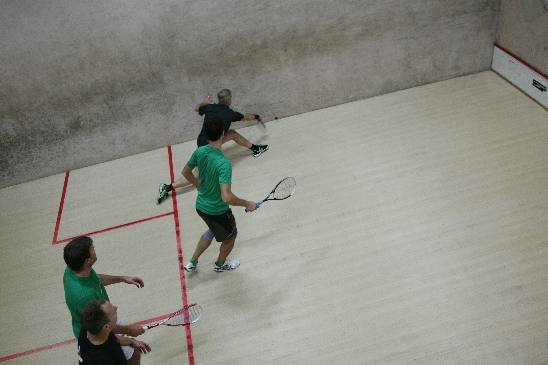}
        \caption*{Normal}
    \end{subfigure}
    \begin{subfigure}[c]{0.26\linewidth}
        \centering
        \includegraphics[height=1.8cm]{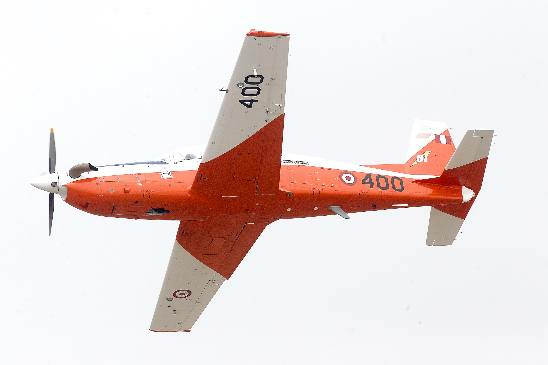}
        \caption*{Bright}
    \end{subfigure}
    \\
    \begin{minipage}[c]{0.15\linewidth}
        \centering
        \textbf{Depictions}
    \end{minipage}
    \begin{subfigure}[c]{0.26\linewidth}
        \centering
        \includegraphics[height=1.8cm]{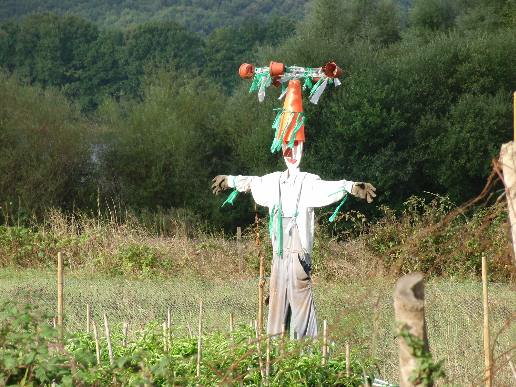}
        \caption*{Person}
    \end{subfigure}
    \begin{subfigure}[c]{0.26\linewidth}
        \centering
        \includegraphics[height=1.8cm]{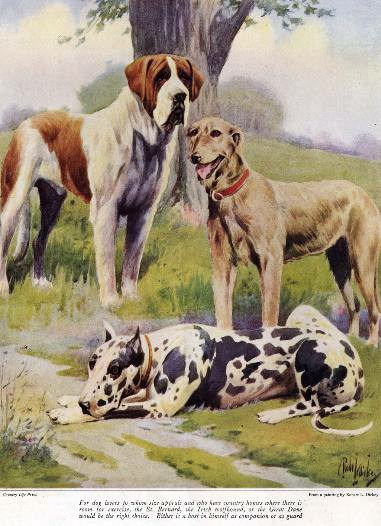}
        \caption*{Non-Person}
    \end{subfigure}
    \begin{subfigure}[c]{0.26\linewidth}
        \centering
        \includegraphics[height=1.8cm]{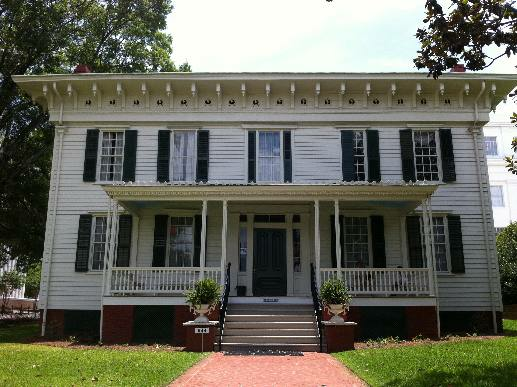}
        \caption*{No Depiction}
    \end{subfigure}
    \caption{Images from each fine-grained benchmark dataset.}
    \label{fig:extensive_fine_grained_benchmark_examples}
\end{figure*}

\subsection{Fine-grained Benchmark Suite}\label{sec:benchmark_suite}
While Wake Vision provides a substantial improvement in scale and quality over existing datasets, we recognize that overall test set performance alone may not capture the nuanced challenges of real-world deployment for TinyML applications. 
Internet-sourced images typically exhibit a distinct bias towards well-composed photographs that are well-lit, properly framed, and feature clearly visible subjects. 
This stands in stark contrast to the challenging conditions where TinyML person detection systems are typically deployed, such as varying lighting conditions, extreme viewing distances, or partial occlusions~\citep{GreatTin73:online}.

To bridge this gap between benchmark performance and practical utility, we augment Wake Vision with a comprehensive fine-grained benchmark suite. 
This suite evaluates model robustness across challenging real-world scenarios where traditional accuracy metrics might mask significant failure modes. 
The benchmarks assess performance across critical dimensions, including lighting conditions, subject distance, and demographic attributes, enabling developers to identify potential biases or limitations during the design phase rather than after deployment. 
The suite comprises five fine-grained benchmark sets, three of which (Distance, Lighting, and Depictions) are applicable to any dataset generated by the Wake Vision pipeline, while the other two (Perceived Gender and Perceived Age) are specific to Wake Vision; \Cref{fig:extensive_fine_grained_benchmark_examples} shows examples from each.

Each of the five fine-grained benchmarks has been picked based on a combination of its relevance to TinyML use cases and the availability of requisite metadata to generate the sets.
Each benchmark set is a subset of the respective validation or test set filtered based on the criteria under test.
The benchmark determines whether a model is sufficiently accurate in the planned deployment setting. For example, a model designer may make different design choices for a use case where the subject is close to the camera and well-lit compared to the inverse setting.
Appendix~\ref{app:case_study} contains a case study that shows how fine-grained evaluation sets can be used to design robust models.

\paragraph{Perceived Gender and Age}

Underrepresented sub-groups typically constitute a small portion of a generic test set; therefore, top-line metrics often obscure a bias in a model until it is deployed~\citep{hooker2019compressed}.
This benchmark, generated from the Open Images \gls{miap} extended fairness labels \citep{miap_aies}, evaluates a model separately on demographics that are underrepresented in the underlying dataset distribution, identifying bias. These labels are based on perceived gender and age representation and are not necessarily representative.

\paragraph{Distance}
This tests how the distance of people in images impacts model performance.
If a person detection system is intended to recognize subjects at great distances, the system's performance on the faraway dataset will be more informative than its performance on the top-level test set.
We created three datasets based on the percentage of the image the subject bounding box covers. The three sets are near (\textgreater{}60\%), at a medium distance (10-60\%), and far away (\textless{}10\%).

\paragraph{Lighting}
The lighting data sets test the performance of \gls{ml} systems in different lighting conditions.
In scenarios like security monitoring, outdoor robotics, or augmented reality applications, models must be robust to varying lighting conditions, including low-light environments. 
We create three fine-grained datasets of this type for dark, normal, and bright lighting conditions, respectively.
We quantify lighting conditions by the average pixel values of images in greyscale, a simple but effective method for distinguishing lighting conditions~\citep{zhang2017research}. We define low as an average pixel value less than 85, normal as between 85 and 170, and bright lighting conditions as greater than 170.

\paragraph{Depictions}
A particularly challenging task for a person detection model is to correctly reject depictions of people.
In many use cases, a person detection model should not trigger on a depiction. 
For example, a room occupancy detector could incorrectly identify a painting on the wall as a person.
This benchmark measures a model's accuracy on three related sets of non-person samples: depictions of people, depictions that are not of people, and images that do not contain a depiction of any kind.
Depictions of people can range from photo-realistic to crude stick figures.

\subsection{Community Accessibility and Involvement}
Wake Vision and its fine-grained benchmark suite are available through TensorFlow Datasets \citep{TFDS} and HuggingFace Datasets~\citep{lhoest-etal-2021-datasets} to enable easy access and use by the community. 
The images and labels are rehosted to ensure the dataset will not shrink due to dead links. 
Removal of images can be requested at the email address reported in Appendix~\ref{app:image-removal}.
The rehosted labels are generated according to our default dataset configuration, further described in Appendix~\ref{app:label_generation_details}.
Most datasets remain static after the initial release. To continuously improve Wake Vision, we partner with the Edge AI Foundation~\citep{edgeaifoundation} to gather community involvement through recurrent competitions (Section~\ref{sec:wake_vision_challenge}). It is our vision that the iterative enhancements brought by the community are integrated into the dataset, leading to a continuous improvement cycle.  

\subsection{\new{Generating Binary Image Classification Datasets for TinyML}}
\label{sec:other_binary_datasets}
Beyond person detection, many other TinyML applications can benefit from large-scale binary classification datasets. 
For example, bird detection enables smart nest systems~\citep{debauche2020smart} and automated feeding ~\citep{gerhardson2022design}, while vehicle detection enables automated parking space occupancy monitoring~\citep{bura2018edge} and blind spot monitoring~\citep{shen2018blind}. In particular, Open Images v7~\citep{OpenImages, OpenImages2} contains 9.6K categories and 600 boxable objects, such as birds and vehicles, significantly exceeding the 80 object categories available in MS COCO~\citep{lin2014microsoft}. In order to provide users with a quick way to bootstrap a dataset and model for an arbitrary TinyML vision task, the Wake Vision pipeline targets any of the classes in Open Images. We provide a comparison between the Wake Vision pipeline and \gls{vww} in Section~\ref{sec:birds_cars_results}.

%% file: sec/4_results.tex
\section{Results}
We present a comprehensive evaluation of the Wake Vision dataset and pipeline through the following analyses.
First, we compare our two training sets to derive insights into best practices when training.
Second, we demonstrate that Wake Vision is an effective drop-in replacement for \acrfull{vww} by conducting cross-evaluation studies using identical MobileNetV2 models.
Third, we validate Wake Vision's benefits across diverse model architectures used in the TinyML community, from MCUNet~\citep{lin2020mcunet} to ColabNAS~\citep{garavagno2024colabnas} families.
Fourth, we present fine-grained benchmark results across demographics, environmental conditions, and visual contexts to assess model robustness in real-world scenarios.
Fifth, we show that the Wake Vision pipeline produces useful binary classification datasets beyond person detection, demonstrating it on bird and car detection.
Sixth, we show models trained on Wake Vision maintain their higher performance compared to those trained on \gls{vww} when evaluated on several Out-of-Distribution (OOD) datasets, such as driving and overhead surveillance imagery.
Seventh, we evaluate the impact of dataset quality and size on low-capacity model training to inform future data-centric efforts.
Eighth, we present the results of the first Wake Vision Challenge and the community-driven improvements it has yielded for the Wake Vision dataset.
We close with lessons that generalize beyond person detection for data-centric TinyML.
Through these analyses, we demonstrate that Wake Vision not only improves upon \gls{vww}'s performance but also enables a more thorough evaluation of TinyML computer vision models.

\subsection{Training Set Evaluation}\label{sec:training-set-eval}
\begin{table}
    \centering
    \caption{Accuracy on test set of a MobileNetV2-0.25 model trained on image-level and bounding box labels, with and without KD. A MobileNetV2-1.0 model trained on the bounding box set is the teacher.
    }
    \label{tab:image-level-bbox-cross-val}
    \resizebox{\columnwidth}{!}{
    \begin{tabular}{lcccc}
        \toprule
        Training set & Label Source & Training Set Size & Accuracy & Distilled Accuracy \\
        \midrule
        Wake Vision (Large) & Image-Level & 5,760,428 & 	80.80$\pm$0.18\% & 85.00$\pm$0.20\% \\
        Wake Vision (Quality) & Bounding Box & 1,248,230 & 84.89$\pm$0.11\% & 85.17$\pm$0.18\% \\
        Wake Vision (Combined) & Both & 5,760,428 & 85.72$\pm$0.04\% & - \\
        \bottomrule
    \end{tabular}
}
\end{table}
To evaluate the performance of the two Wake Vision training sets, we train identical models on each set.
We use MobileNetV2~\citep{sandler2018mobilenetv2} models with a width modifier of 0.25 for 200,000 steps on 224x224x3 images using AdamW~\citep{loshchilov2017decoupled},  a learning rate of 2e-3 with cosine decay and a weight decay of 4e-6.
After training, we compare the performance of each model on the common test set.

The result of this comparison is shown in~\cref{tab:image-level-bbox-cross-val}. 
The Wake Vision (Quality) training set outperforms the Wake Vision (Large) training set by 4.09\% test accuracy, indicating that label quality is more important than quantity in this evaluation setting.
The gap between the two is reduced to just 0.17\% when training exclusively on soft labels from a teacher model (MobileNetV2-1.0) trained on bounding box labels. 
We further study how training set size and quality impact accuracy across model sizes, and observe that smaller models are more sensitive to errors in the training set, dropping accuracy by a delta of 0.8\% compared to larger models (\cref{sec:quality_vs_size}).

Using the two training sets in unison to train a model where Wake Vision (Large) acts as a pre-training set, and Wake Vision (Quality) acts as a fine-tuning set achieves our best performance at a mean top-level test accuracy of 85.72\%. 
The two training sets together provide a foundation for further research on data-centric TinyML~\citep{mazumder2024dataperf, njor2023data}.

\subsection{Wake Vision \& VWW Cross Evaluation}
\label{sec:comparison_to_vww}

\begin{table}[t]
    \centering
    \caption{Accuracy on the Wake Vision and \gls{vww} test sets by models trained on the \gls{vww}, Wake Vision (Quality), and Wake Vision (Combined) training sets.}
    \label{tab:wv-vww-cross-val}
    \setlength{\tabcolsep}{6pt}
    \small
    \begin{tabular}{ccccc}\toprule
        & & \multicolumn{2}{c}{Train}\\
        & & \gls{vww} & Wake Vision (Quality) & Wake Vision (Combined) \\
        \multirow{2}{*}{\rotatebox[origin=c]{90}{Test}} &  \gls{vww} &88.33$\pm$0.29\% & \textbf{88.59$\pm$0.17\%} & \textbf{89.34$\pm$0.02\%} \\
        & Wake Vision &83.79$\pm$0.23\% & \textbf{84.89$\pm$0.11\%} & \textbf{85.72$\pm$0.04\%}\\\bottomrule
    \end{tabular}
\end{table}

A key challenge in advancing TinyML research is ensuring that improvements can be readily adopted by the community without requiring significant modifications to existing workflows or architectures. 
While novel datasets can offer better quality or scale, their practical impact is limited if they require substantial changes to model architectures, training pipelines, or deployment processes. 
We therefore designed Wake Vision to serve as a drop-in replacement for VWW, and validate this drop-in capability through a cross-evaluation study.
We train two identical MobileNetV2~\citep{sandler2018mobilenetv2} models using the same training recipe from \cref{sec:training-set-eval}.
One model uses \gls{vww}'s training set, and the other uses Wake Vision's Quality training set.
After training these identical models using the respective datasets' training sets with the same recipe for an equal number of steps, we evaluate their performance on the corresponding test sets.

\Cref{tab:wv-vww-cross-val} shows the result of the cross-evaluation. 
The Wake Vision trained model outperforms the \gls{vww} trained model with a 0.26\% improvement on \gls{vww}'s own test set, which indicates that Wake Vision is a direct improvement over \gls{vww}, not simply a domain shift. 
We also achieve a 1.1\% improvement over \gls{vww} on Wake Vision's Test set, indicating that the new test set is more challenging.
\Cref{tab:wv-vww-cross-val} also shows the performance of the Wake Vision (Combined) model described in \cref{sec:labeling}.
This model achieves 0.75\% improvement over the Wake Vision (Quality) trained model on the \gls{vww} test set and 0.83\% improvement on the Wake Vision test set.
The Wake Vision (Combined) model is trained for longer than the \gls{vww} and Wake Vision (Quality) models but shows the value of using the training sets in unison even on the \gls{vww} test set.

\subsection{Evaluation Across TinyML Model Architectures}\label{sec:comprehensive_model_architecture_evaluation}
To show that the benefits of using Wake Vision over \gls{vww} are not limited to MobileNetV2 architectures, we also train several other models on both training sets, Wake Vision (Quality) and \gls{vww}, and compare their performance.
For this experiment, we use a collection of models commonly used in the TinyML community, such as models from the MCUNet~\citep{lin2020mcunet} and MicroNets~\citep{banbury2021micronets} families. 
Furthermore, we include models from ColabNAS~\citep{garavagno2024colabnas} to provide evaluation results on even smaller models.

\begin{figure}[t]
    \centering    
    \subcaptionbox{Initial Pareto frontier for Wake Vision vs \gls{vww}\label{fig:cross-eval-wv}}{\includegraphics[width=0.495\linewidth]{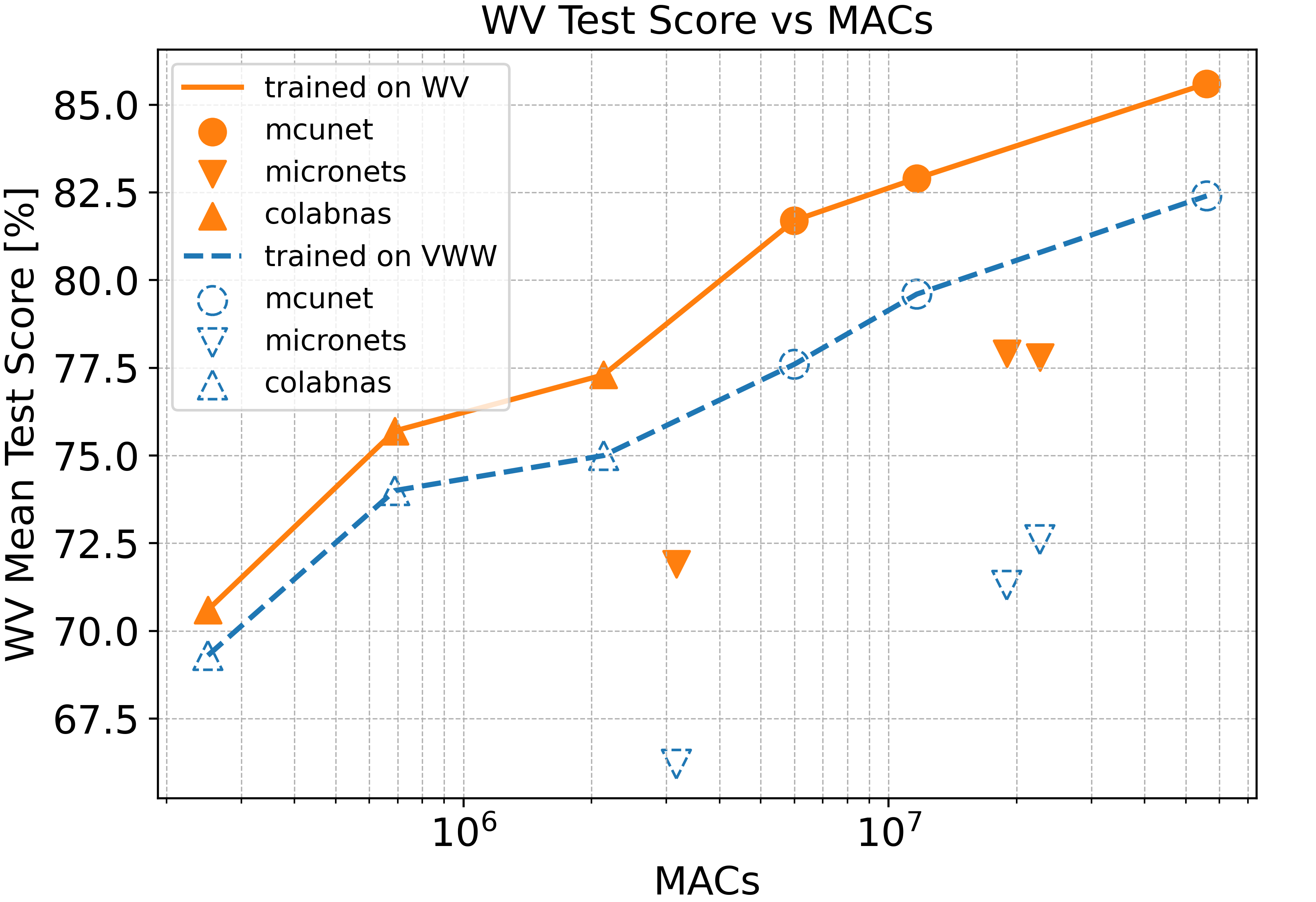}}
    \hfill
    \subcaptionbox{Advanced Pareto frontier by challenge submissions\label{fig:pareto_frontier_after_challenge}}{\includegraphics[width=0.495\linewidth]{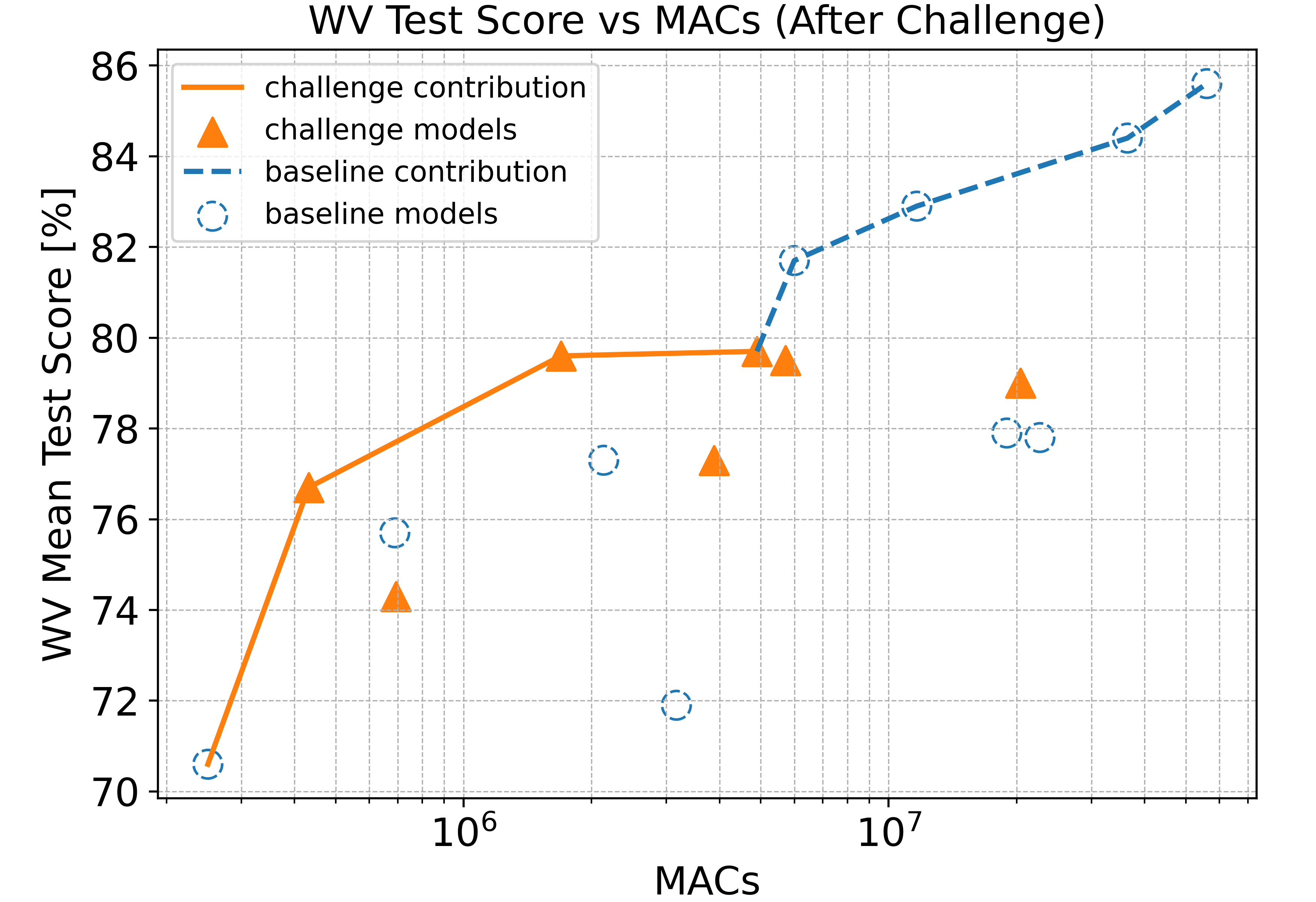}}
    \caption{The Wake Vision Challenge submissions advanced our initial Pareto frontiers.}
\end{figure}

Each model is trained on each dataset using the Adam optimizer with a learning rate of 1e-3 for 243,800 steps at a batch size of 512, equivalent to 100 epochs on the Wake Vision (Quality) training set. 
In addition, we apply random horizontal flips, rotations and use 8-bit quantization-aware training.
We use the checkpoint with the best validation score for testing.
A desktop computer featuring a 13th Gen Intel® Core™ i9-13900K × 32, 32 GB of RAM, and an NVIDIA GeForce RTX 4080 with 16 GB of VRAM has been employed to train the TinyML models.

We cross-evaluate each trained model's performance on the test sets of both \gls{vww} and Wake Vision.
Results on the Wake Vision test set are presented in~\cref{fig:cross-eval-wv}, while further details and the results for the \gls{vww} test set can be found in Appendix~\ref{app:additional_benchmark_results}.
For the majority of models, the one trained on the Wake Vision (Quality) set achieves a higher accuracy on both the \gls{vww} and Wake Vision test set, which supports our claim that Wake Vision is a direct improvement over \gls{vww}.
At the extreme, for one MicroNets~\citep{banbury2021micronets} model (middle right of \cref{fig:cross-eval-wv}), switching the training set to Wake Vision increased the test accuracy by as much as 6.6\%.
The only instance where Wake Vision isn't a complete improvement over \gls{vww} is in \cref{fig:cross-eval-vww} in Appendix~\ref{app:additional_benchmark_results}, where small models perform better when trained on the same dataset they are tested on (i.e., trained on \gls{vww} and tested on \gls{vww}), likely due to ultra-low-capacity models failing to generalize past the slight domain shift between the \gls{vww} and Wake Vision test sets.
Accuracy and performance metrics, such as latency or model size, are in constant tension in TinyML models, so any boost in accuracy achieved through Wake Vision can also be traded off for more efficient architectures (in some cases 3$\times$ fewer MACs) while preserving test accuracy.

\subsection{Fine-Grained Benchmark Evaluation}\label{sec:fine_grained_datasets.benchmark_results}

Aggregate metrics like accuracy or F1-score provide high-level insight but can mask critical failure modes and biases that emerge in real-world scenarios. 
In response, we include fine-grained benchmarks in Wake Vision (Section~\ref{sec:benchmark_suite}) across demographics, environmental conditions, and visual contexts. 
This is particularly important for person detection models deployed in real-world applications, where failures on certain subgroups or conditions could adversely affect fairness and safety. 

In Table~\ref{tab:benchmark_suite_transposed_compact}, we compare the models from \cref{sec:comparison_to_vww}, trained on Wake Vision and \gls{vww}, across various scenarios to understand their relative strengths and limitations. 
The Wake Vision model exhibits superior robustness across the challenging settings exercised by our benchmarking suite. 
For instance, on the ``Depictions'' benchmark, which evaluates performance on images containing persons, non-person objects, or no depictions, the Wake Vision model achieves an F1 score of 0.71 for person depictions, outperforming the \gls{vww} model's 0.66. 
The Wake Vision model also showcases improved performance on the ``Age'' benchmark, with F1 scores of 0.94, 0.91, and 0.94 for young, middle, and older individuals, surpassing the \gls{vww} model's scores of 0.90, 0.88, and 0.89, respectively. 
This highlights the dataset's effectiveness in enhancing model robustness for detecting people across various age groups, particularly the elderly demographic. 
In addition, the Wake Vision model demonstrates resilience to challenging lighting conditions, achieving F1 scores of 0.85 in dark lighting scenarios, compared to the \gls{vww}'s 0.81.

Notably, a \gls{vww} model only significantly outperforms a Wake Vision model on the far distance benchmark at a score of 0.67 and 0.59, respectively.
During dataset generation, \Gls{vww} filters out persons smaller than 0.5\% of the image, versus 5\% for Wake Vision. 
Therefore, \gls{vww} contains images with significantly smaller persons.
We did experiment with adopting a default of 0.5\% for Wake Vision, however this disproportionately decreased the performance of models in other settings.

\begin{table}[t]
\centering
\caption{Wake Vision Fine-Grained Benchmark Suite. We report the samples in each set and the average F1 score across three Wake Vision models and three \gls{vww} models on each benchmark.}
\label{tab:benchmark_suite_transposed_compact}
{
\tiny 
\setlength{\tabcolsep}{3pt} 

    \begin{tabular}{ll|rrr|rrrr|rrr|rrr|rrr}
        \toprule
        \multicolumn{2}{c|}{} & \multicolumn{3}{c|}{Gender} & \multicolumn{4}{c|}{Age} & \multicolumn{3}{c|}{Distance} & \multicolumn{3}{c|}{Lighting} & \multicolumn{3}{c}{Depictions} \\
        \cmidrule(lr){3-5} \cmidrule(lr){6-9} \cmidrule(lr){10-12} \cmidrule(lr){13-15} \cmidrule(lr){16-18}
        \multicolumn{2}{c|}{} & \rotatebox[origin=c]{295}{Female} & \rotatebox[origin=c]{295}{Male} & \rotatebox[origin=c]{295}{Unknown} & \rotatebox[origin=c]{295}{Young} & \rotatebox[origin=c]{295}{Middle} & \rotatebox[origin=c]{295}{Older} & \rotatebox[origin=c]{295}{Unknown} & \rotatebox[origin=c]{295}{Near} & \rotatebox[origin=c]{295}{Medium} & \rotatebox[origin=c]{295}{Far} & \rotatebox[origin=c]{295}{Dark} & \rotatebox[origin=c]{295}{Normal} & \rotatebox[origin=c]{295}{Bright} & \rotatebox[origin=c]{295}{Person} & \rotatebox[origin=c]{295}{Non-Person} & \rotatebox[origin=c]{295}{No Depiction} \\
        \midrule 

        \multirow{2}{*}{\rotatebox[origin=c]{90}{Size}} & Val & 684 & 1310 & 1612 & 275 & 2133 & 90 & 1299 & 5457 & 2213 & 398 & 3255 & 14315 & 1012 & 356 & 352 & 8583 \\
        & Test & 2157 & 3918 & 4940 & 884 & 6595 & 276 & 3837 & 16333 & 6876 & 1140 & 9420 & 43010 & 3332 & 978 & 1101 & 25802 \\
        \midrule 

        \multirow{2}{*}{\rotatebox[origin=c]{90}{F1}} & Wake Vision & \bfseries 0.93 & \bfseries 0.91 & 0.77 & \bfseries 0.94 & \bfseries 0.91 & \bfseries 0.94 & 0.71 & \bfseries 0.91 & \bfseries 0.85 & 0.59 & \bfseries 0.85 & \bfseries 0.82 & \bfseries 0.82 & \bfseries 0.71 & \bfseries 0.86 & \bfseries 0.87 \\
        & \gls{vww} & 0.89 & 0.88 & \bfseries 0.78 & 0.90 & 0.88 & 0.89 & \bfseries 0.74 & 0.89 & 0.84 & \bfseries 0.67 & 0.81 & \bfseries 0.82 & \bfseries 0.82 & 0.66 & 0.82 & 0.85 \\
        \bottomrule
    \end{tabular}
}
\end{table}
Across the five fine-grained axes, Wake Vision-trained models match or exceed VWW-trained models on 13 of 16 subsets and show the largest gains where TinyML deployment conditions diverge most from typical training imagery: depicted persons, older individuals, and dark lighting.

\subsection{Generalized Binary Image Classification Datasets} \label{sec:birds_cars_results}
We demonstrate that the Wake Vision pipeline can be used for constructing binary image classification datasets for any category of interest among the 9.6K trainable classes or 600 boxable objects present in Open Images v7~\citep{OpenImages, OpenImages2}. 
Datasets constructed by the Wake Vision pipeline are significantly larger than those produced by the prior \gls{vww} methodology applied to the same class. 
In particular, we compare the Wake Vision pipeline against \gls{vww} on \textit{birds} and \textit{cars}, as these two categories are shared across both Open Images and the 80 object categories available in MS COCO.
As noted in Section~\ref{sec:other_binary_datasets}, bird detection has applications in smart nests~\citep{debauche2020smart} and automated feeders~\citep{gerhardson2022design}, while car detection applies to parking monitoring~\citep{bura2018edge} and blind-spot systems~\citep{shen2018blind}.
The car detection dataset generated via the Wake Vision pipeline contains $12\times$ more images than \gls{vww}, with a decrease in label error rate from 6.4\% to 4.8\%.
Similarly, the bird detection dataset provides $27\times$ more images with a decrease from 6.6\% to 0.6\% in label error rate.

\begin{figure}[t]
    \centering
    \begin{subfigure}[b]{0.45\textwidth}
        \centering
        \includegraphics[width=\textwidth]{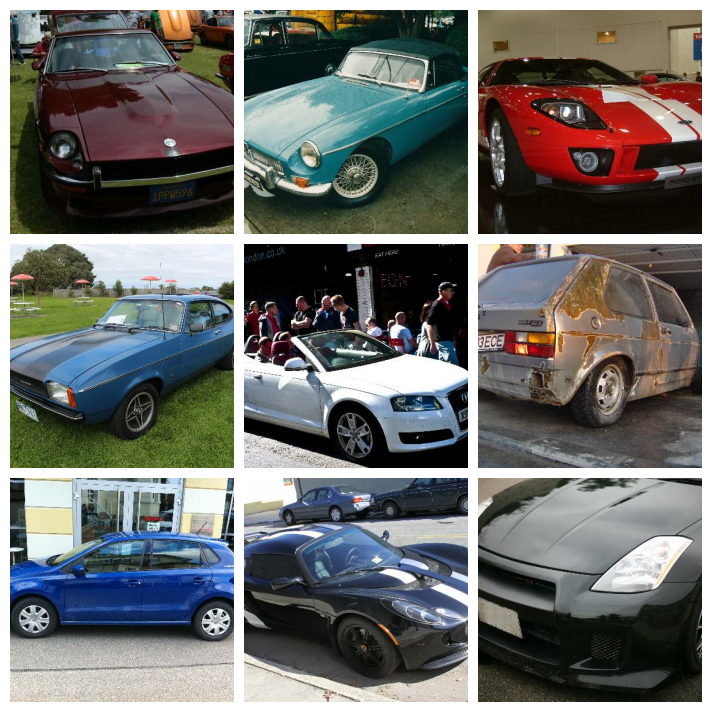}
        \caption{Wake Vision ``Car'' samples}
        \label{fig:wv_car_examples}
    \end{subfigure}
    \hfill
    \begin{subfigure}[b]{0.45\textwidth}
        \centering
        \includegraphics[width=\textwidth]{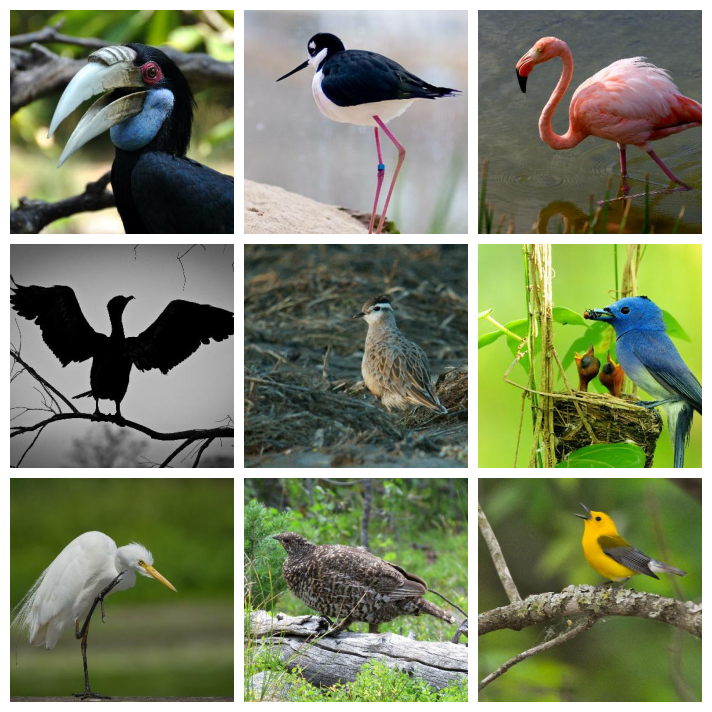}
        \caption{Wake Vision ``Bird'' samples}
        \label{fig:wv_bird_examples}
    \end{subfigure}
    \caption{Positive class examples from the test splits of our Wake Vision generated datasets for ``Car'' and ``Bird'' objects, demonstrating our ability to easily prototype large-scale TinyML vision datasets (\Cref{sec:other_binary_datasets}).}
    \label{fig:car_bird}
\end{figure}

Table \ref{tab:car_and_bird} compares the number of images as well as the label error rate for the car and bird datasets, when generated using both our methodology and the \gls{vww} one. All datasets are balanced, i.e., have the same number of images for the target and background class. Label error rates have been manually computed on a subset of 500 randomly selected images for each generated dataset. Examples of positive class images from the Wake Vision test splits are shown in \Cref{fig:car_bird}.

\noindent\begin{table}[t!]    
    \centering
    \caption{Sample Counts of Car and Bird datasets.}
    \label{tab:car_and_bird}
    \begin{tabular}{c c c} 
        \toprule
        \bf{Methodology} & \bf{\# of Images} & \bf{Label Error Rate}  \tabularnewline \midrule
        \multicolumn{3}{c}{\bf{Car}} \tabularnewline 
        \bf{Wake Vision (Ours)}    & 157,668          & 4.8\% \tabularnewline    
        VWW~\citep{chowdhery2019visual}              & 12,902           & 6.4\% \tabularnewline \midrule
        \multicolumn{3}{c}{\bf{Bird}} \tabularnewline 
        \bf{Wake Vision (Ours)}    & 31,642           & 0.6\% \tabularnewline 
        VWW~\citep{chowdhery2019visual}              & 1,148            & 6.6\% \tabularnewline       
        \bottomrule
        \end{tabular}
\end{table}

\subsection{\new{Accuracy under Data Distribution Shifts}}\label{sec:ood}

\begin{figure}[t]
    \centering
    \subcaptionbox{TUD-MotionPairs\label{fig:tudmp}}{
        \includegraphics[height=3.5cm]{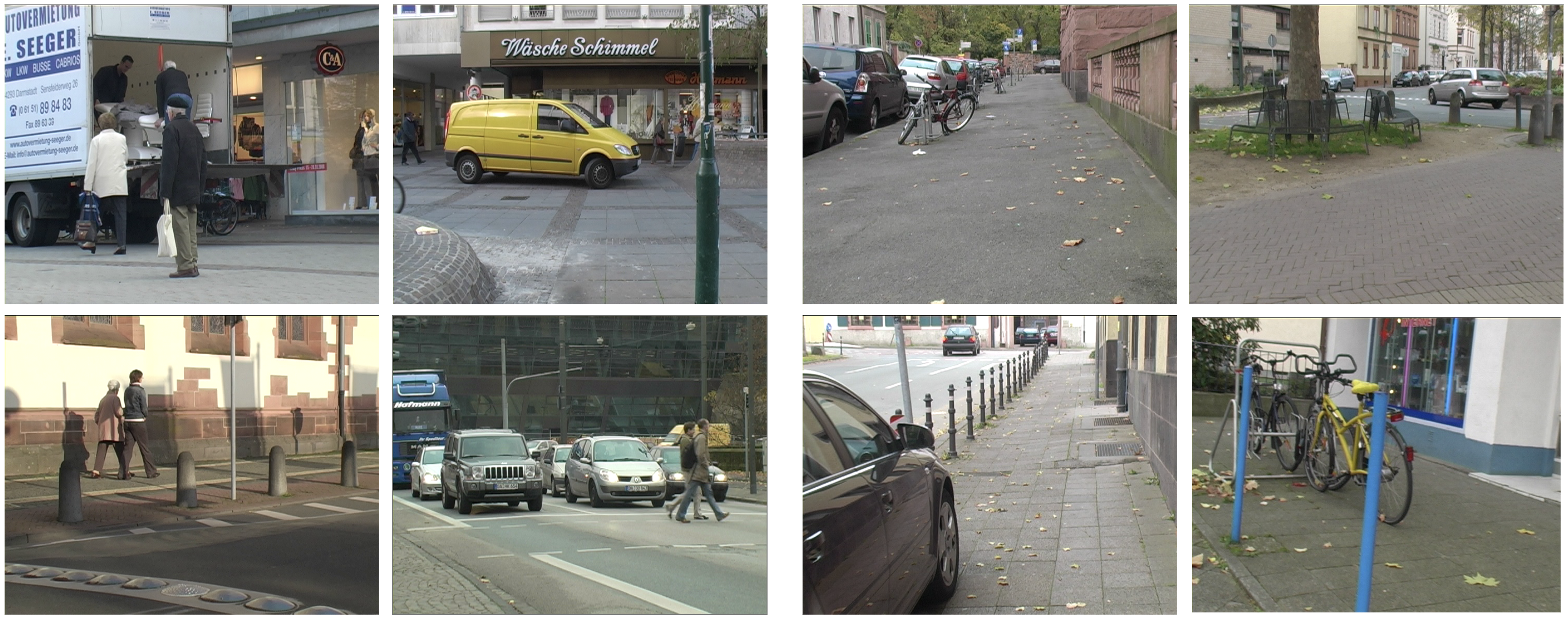}
    }

    \par\vspace{0.5em}

    \subcaptionbox{Kaggle1\label{fig:kaggle1}}{
        \includegraphics[height=3.5cm]{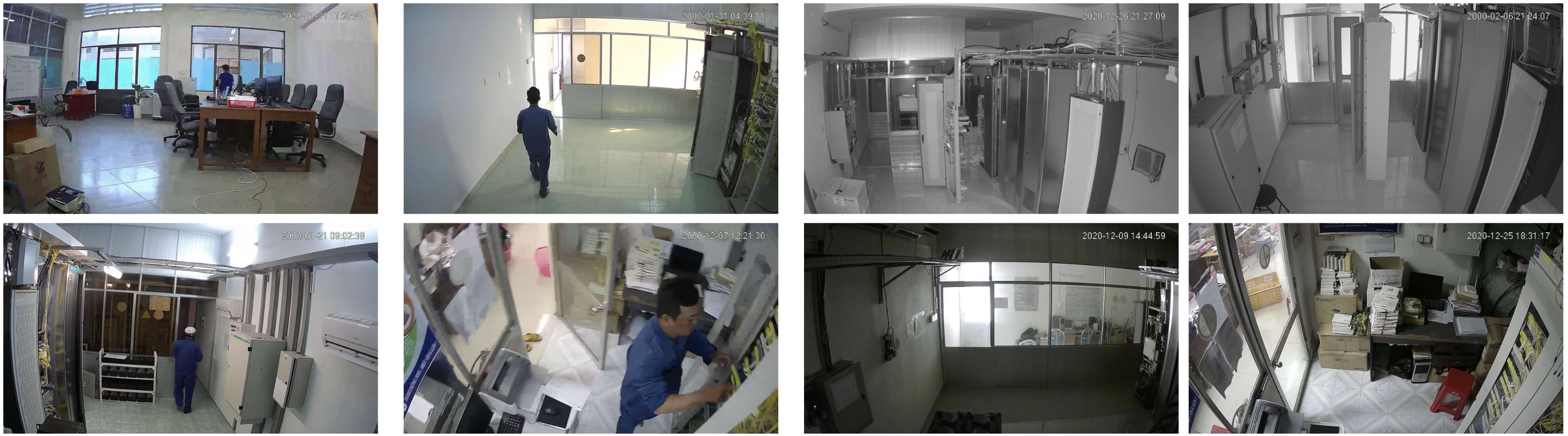}
    }

    \par\vspace{0.5em}

    \subcaptionbox{Kaggle2\label{fig:kaggle2}}{
        \includegraphics[height=3.5cm]{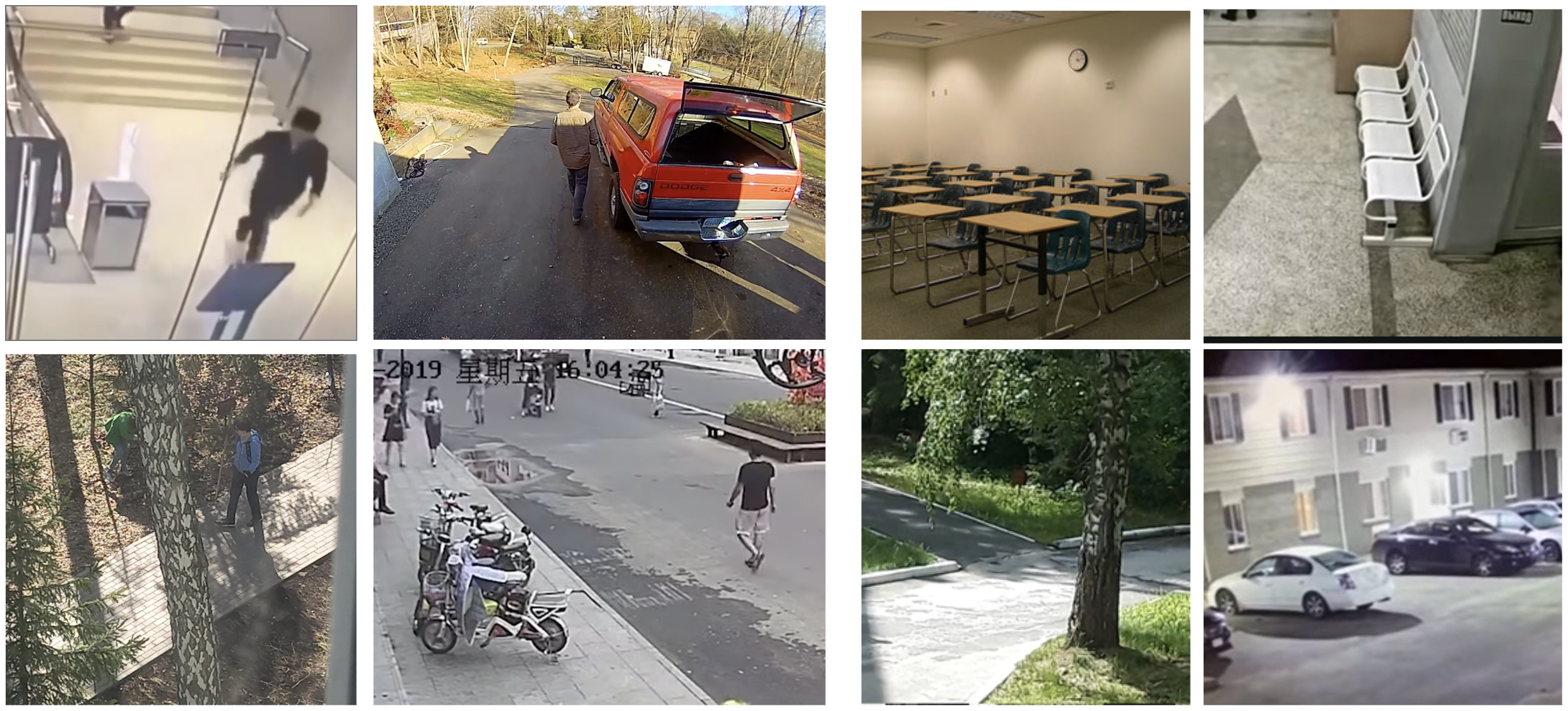}
    }

    \caption{Four positive and four negative examples drawn randomly from each of our three Out-of-Distribution datasets.}\label{fig:ood-examples}
\end{figure}

\begin{figure}[t]
    \centering    
    \subcaptionbox{TUD-MotionPairs\label{fig:pareto-tudmp}}{\includegraphics[width=0.31\linewidth]{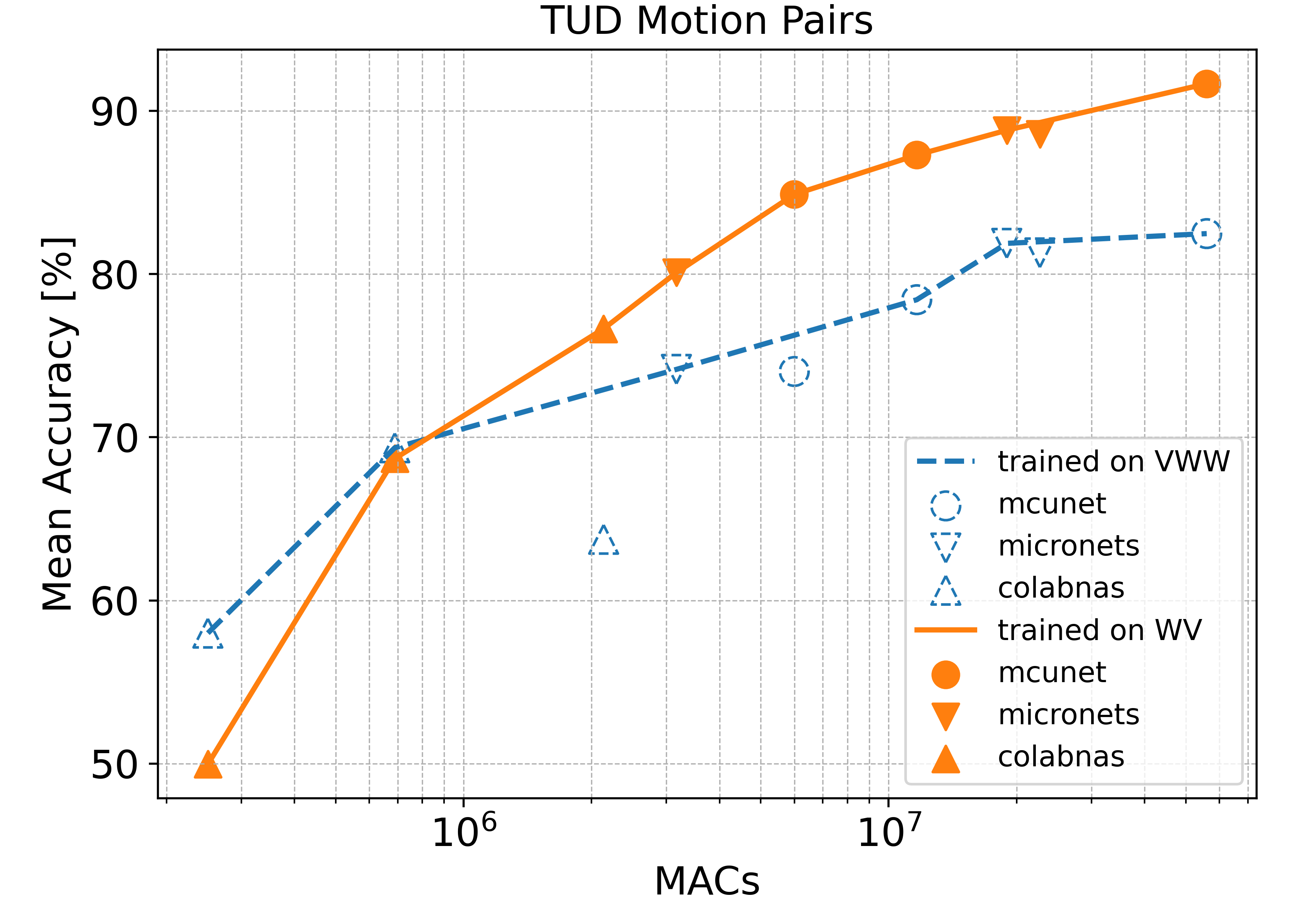}}
    \hfill
    \subcaptionbox{Kaggle1\label{fig:pareto-kaggle1}}{\includegraphics[width=0.31\linewidth]{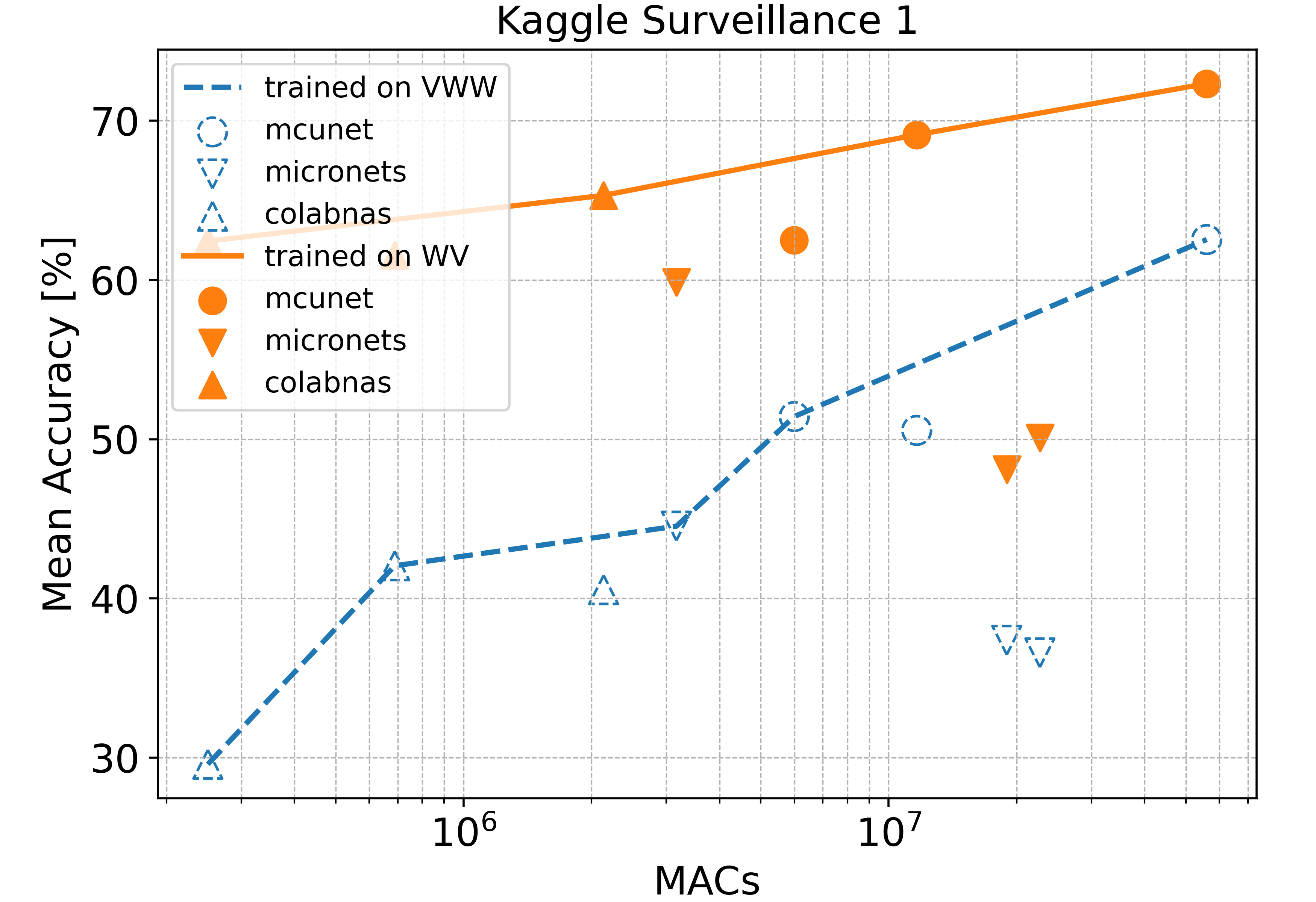}}
    \hfill
    \subcaptionbox{Kaggle2\label{fig:pareto-kaggle2}}{\includegraphics[width=0.31\linewidth]{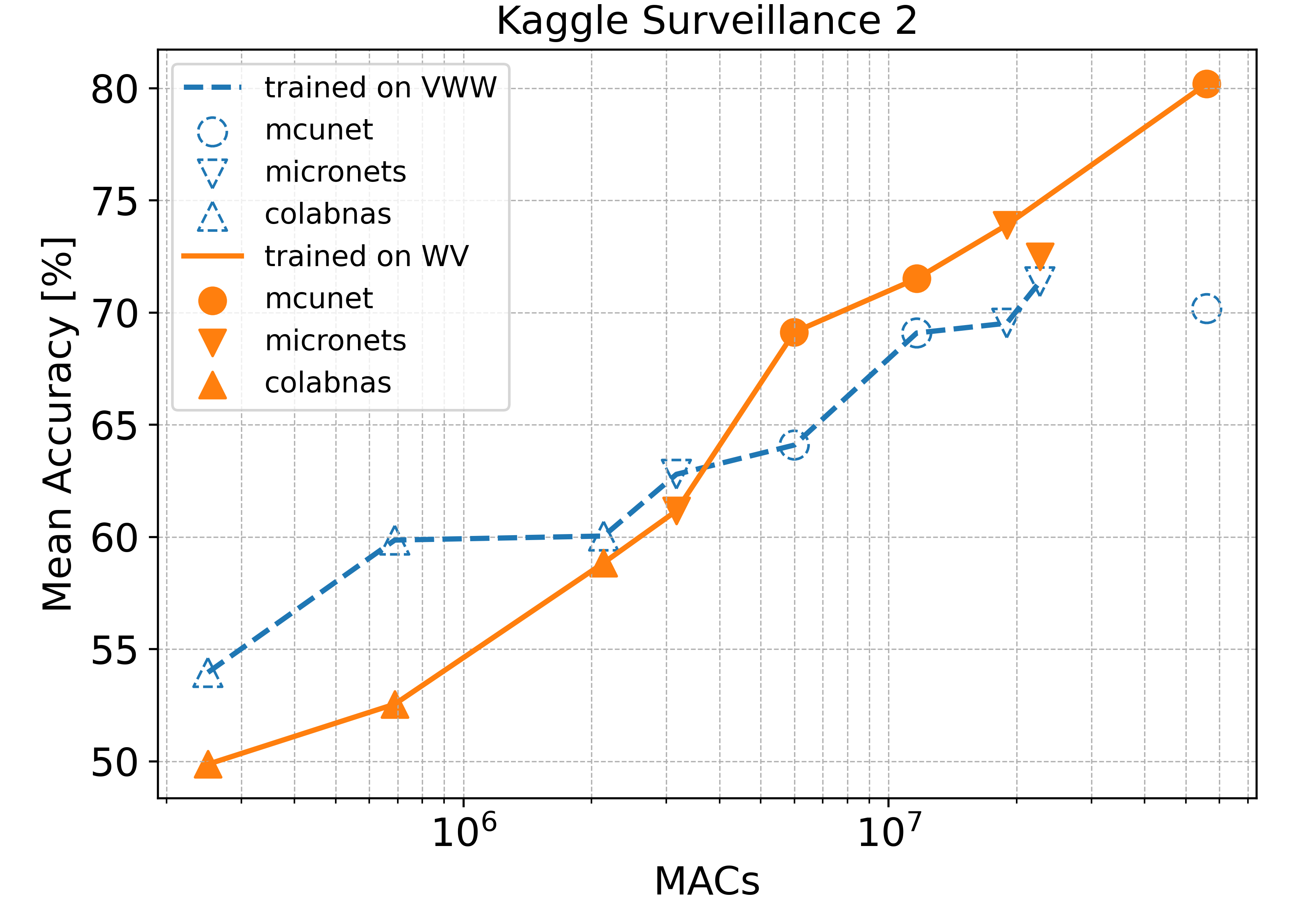}}
    \caption{Pareto frontiers for Wake Vision vs \gls{vww} on Out-of-Distribution datasets}\label{pareto-ood}
\end{figure}

In order to assess the generalizability of Wake Vision to other data distributions, we simulate data distribution shifts~\citep{beery2018recognition, taori2020measuring} by running inference on a selection of three new datasets using the models already trained on Wake Vision and \gls{vww} from Section~\ref{sec:comprehensive_model_architecture_evaluation} and computing Pareto curves on these new datasets. In particular, we seek to emulate downstream applications of person detection (such as CCTV surveillance) when using Wake Vision for training data. The upstream data sources of Open Images for Wake Vision and MS-COCO for \gls{vww} exhibit image characteristics common in handheld photographs, often with center-biased locations of nearby subjects that are large in frame and with minimal barrel distortion. We wish to evaluate how Wake Vision performs under data distribution shifts, such as with surveillance imagery which is often taken from an overhead perspective in challenging lighting conditions and in the presence of fisheye lens distortion.  

We select the TUD-MotionPairs dataset \citep{wojek2009multi} (2,185 positives and 385 negatives) containing imagery from a vehicle's driving perspective to emulate pedestrian or cyclist detection, and two Kaggle datasets containing surveillance imagery from distant and overhead perspectives: Surveillance Images for Person Detection \citep{martins2025surveillance} (5,265 positives and 1,338 negatives) and Human Detection Dataset \citep{werner2025human} (559 positives and 362 negatives). Randomly selected examples from each dataset are shown in Figure~\ref{fig:ood-examples}. Due to their unusual perspective, lighting, and lens properties, these three datasets could be considered out-of-distribution (OOD) with respect to most images in Open Images and MS-COCO used when training the models in Section~\ref{sec:comprehensive_model_architecture_evaluation}. We evaluate these pretrained models on our three selected OOD datasets to compare the performance of Wake Vision-trained models against \gls{vww}-trained models under data distribution shift at inference time. Figure~\ref{pareto-ood} presents our results. Models trained on Wake Vision generally outperform models trained on \gls{vww} under these distribution shifts.
This suggests that Wake Vision's larger scale and wider diversity of image characteristics improve generalization, though the size of the advantage varies with the target distribution.


\begin{figure*}[t]
    \centering
    \includegraphics[width=\linewidth]{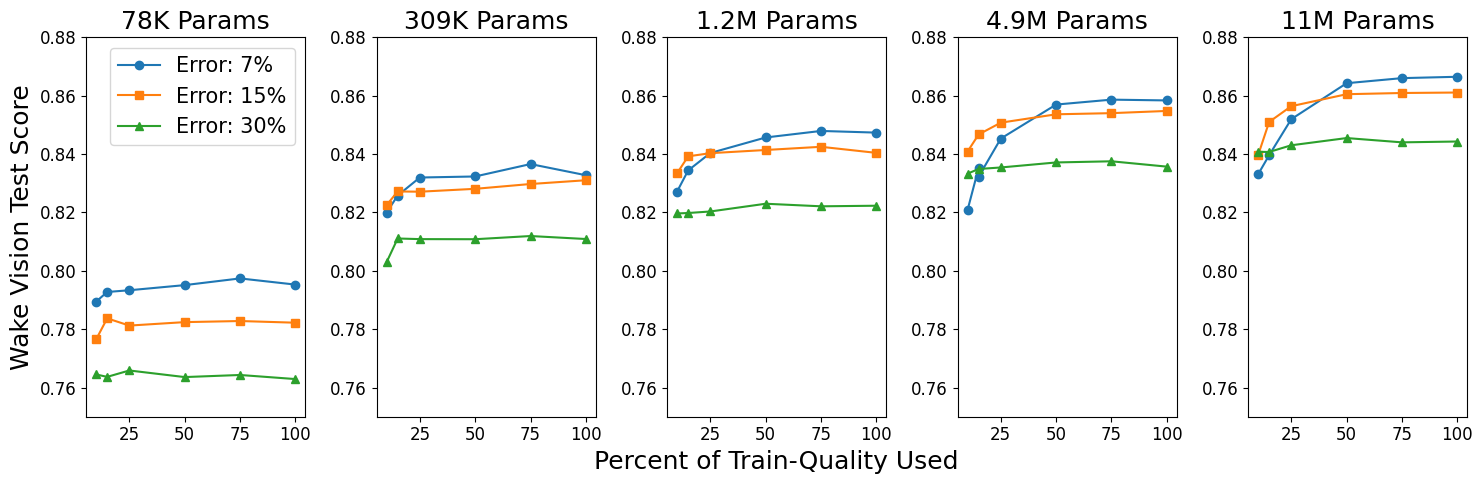}
    \caption{Impact of dataset error rate and size on models of varying capacity.
    The $y$-axis is the Wake Vision test accuracy and the $x$-axis is the percentage of Wake Vision Train (Quality) used in training.
    The error \% indicates the expected rate of incorrectly labeled samples; the 7\% line corresponds to the baseline Wake Vision (Quality) error rate of approximately 6.8\% (Table~\ref{tab:wv-vww-test-errors}).}
    \label{fig:error_vs_dataset_size}
\end{figure*}

\subsection{Impact of Dataset Quality \& Size in TinyML}
\label{sec:quality_vs_size}

Understanding the relationship between dataset quality and size is important for training efficient models in real-world scenarios.
While gathering large amounts of data has become easier, ensuring high-quality labeling remains costly and time-consuming. 
This creates an important trade-off: is it better to have a smaller dataset with high-quality labels or a larger dataset with more label noise? 
The answer likely depends on the capacity of the model and the specific task. 

We investigate the impact of dataset quality and size on Resnet~\citep{he2016deep} style models of varying capacity. 
We measure quality by the approximate rate of label errors. The Wake Vision (Quality) training set has an estimated error rate of approximately 6.8\% (Table~\ref{tab:wv-vww-test-errors}). We simulate higher error rate datasets (15\% and 30\%) by flipping the binary label with a certain probability. Appendix \ref{app:label_error_proof} gives the derivation for the flip probability.
We also sweep the dataset size by taking a slice of Wake Vision (Quality).
Each model is trained for 50,000 steps on 224x224x3 images using AdamW~\citep{loshchilov2017decoupled}, a learning rate of 0.001 with a cosine schedule and a weight decay of 0.004.

\Cref{fig:error_vs_dataset_size} shows that smaller models (leftmost figure) are more sensitive to higher error rates than large models\rv{, similar to prior findings that model capacity can help memorize and ignore errors and outliers~\citep{zhang2021understanding, feldman2020does, goyal2024scaling}}. 
The lowest accuracy drop in the smallest model is 1.3\%, going from the base train (quality) error rate of approximately 6.8\% to 15\% compared to only a 0.5\% accuracy loss for the largest model. 
In contrast, large models benefit more from big datasets and start to overfit on the smaller slices of the training data\rv{~\citep{kaplan2020scaling}}. 
We also see the large slice of the 15\% error rate dataset outperforming a smaller slice of the 6.8\% error rate dataset.

\subsection{\new{The Wake Vision Flywheel: Community-Driven Continuous Improvement}}\label{sec:wake_vision_challenge}
Data-centric TinyML treats the dataset, not just the model, as a maintained artifact. 
Yet most TinyML datasets are released once and then frozen, so label errors, distribution gaps, and missing edge cases discovered post-release have nowhere to go.
Wake Vision instead employs a community-driven flywheel in which competitions surface targeted dataset and model improvements that the maintainer integrates into subsequent releases, turning the dataset into a renewable input for data-centric TinyML research rather than a static benchmark.
We partner with the Edge AI Foundation~\citep{edgeaifoundation} to host Wake Vision competitions.
The competitions held in 2025 focused on two tracks, one on developing high-performance, resource-efficient models, and the other on enhancing the Wake Vision person detection dataset, garnering significant support from the TinyML community with 67 participants across both tracks~\citep{edgeai_wake_vision, edgeai_wake_vision_2}.


The first round produced two concrete data-centric improvements.
On the model track, three new submissions advanced the Wake Vision Pareto frontier (orange triangles in \cref{fig:pareto_frontier_after_challenge}), tightening the accuracy-versus-cost trade-off available to TinyML practitioners on this dataset.
On the data track, a participant's label-correction technique~\citep{WakeVisionChallenge2025_Bedioui}, applied to the Wake Vision (Large) split, reduced its estimated label error rate from 15.2\% to 9.8\% in our experiments (Appendix~\ref{app:data_centric_competition_experiments}), a substantial single-round improvement to a property that conventionally requires expensive manual relabeling.
The implication generalizes beyond person detection. 
Once a dataset is positioned as a maintained artifact, even a small community can supply data-centric improvements that the original authors would not produce alone, and at a cost orders of magnitude below manual relabeling.
New competitions are being developed for the upcoming year, including integration with machine learning educational projects such as MLSys Book~\citep{reddi2024mlsysbook} to engage a new generation of engineers and to keep the data-centric flywheel turning.

\subsection{Lessons for Data-centric TinyML}\label{sec:lessons}

The Wake Vision results yield four lessons that generalize beyond person detection and that we believe apply broadly to data-centric TinyML.

First, \textbf{small models are disproportionately sensitive to training-set label noise}. In our injection protocol (\Cref{sec:quality_vs_size}), the smallest model loses 1.3\% accuracy when training error rises from 6.8\% to 15\%, versus 0.5\% for the largest. When the deployment target is small, label-quality returns dominate scale returns earlier than work on large models suggests.

Second, \textbf{two-stage training combines the strengths of a noisy-large set and a clean-small set}. Pretraining on Wake Vision (Large) and fine-tuning on Wake Vision (Quality) reaches 85.72\% test accuracy, versus 84.89\% for Quality alone and 80.80\% for Large alone (\Cref{sec:training-set-eval}).

Third, \textbf{the manual-labeling budget is best spent where it most affects reported accuracy, not uniformly across the dataset}. Relabeling approximately 70K validation and test images cost \$7,000, while relabeling a 6M training set would cost roughly \$600,000 (\Cref{sec:label_correction}). For data-centric pipelines at scale, the practical question is not whether to manually relabel but where each labeled sample has the largest effect.

Fourth, \textbf{community competitions are a credible mechanism for continuous dataset improvement}. A single round of the Wake Vision Challenge produced a label-correction technique that reduced the Wake Vision (Large) label error rate from 15.2\% to 9.8\% in our experiments (\Cref{sec:wake_vision_challenge}). The cost economics that drive Lesson 3 also apply here. When manual relabeling at training-set scale is prohibitive, sourcing improvements from a community of practitioners is one of the few mechanisms that can keep dataset quality moving in the right direction over time.



%% file: sec/5_ethics.tex
\section{Broader Impact Statement}\label{sec:ethical-considerations}
Releasing a large-scale person detection dataset, a benchmark suite that includes demographic axes, and the Wake Vision pipeline that lowers the cost of building further such datasets carries responsibilities that the empirical results above do not address. We outline these by category below.

\begin{description}
    \item[Potential Misuse:] The Wake Vision pipeline, dataset, and benchmark suite are designed to advance TinyML research and applications. 
    Intended benefits include occupancy detection systems that conserve energy by turning off the lights/HVAC/TV when a person leaves, and privacy preservation through on-device inference, avoiding cloud processing of camera data.
    We acknowledge that person detection technology can be misappropriated for harmful purposes such as weapons targeting or mass surveillance systems.
    \item[Data Rights and Privacy:] Images are sourced from Flickr through Open Images under the CC-BY 2.0 license. 
    Some images may have been uploaded without proper distribution rights or may contain potentially identifying biometric information. 
    Removal of images can be requested at the email address reported in Appendix~\ref{app:image-removal}.
    Due to the scale of the dataset, exhaustive manual verification for offensive content or privacy concerns is infeasible.
    \item[Fairness and Representation:] Our benchmark suite enables disaggregated evaluation across perceived demographics and deployment conditions, helping developers detect failure modes that aggregate accuracy obscures. Disaggregated evaluation does not by itself guarantee equitable deployment, and the suite should be read as a diagnostic tool rather than a certification of fairness; it inherits the demographic categories and limitations of MIAP~\citep{miap_aies}, where attributes are externally perceived rather than self-reported.
    Demographic benchmarks can also be misused to classify gender and age without consent, potentially enabling privacy violations or discriminatory practices. We release the suite for diagnostic evaluation of person detection models, not as endorsement of deploying demographic classifiers.
\end{description}

%% file: sec/6_conclusion.tex
\section{Conclusion}\label{sec:conclusions}

We presented the Wake Vision pipeline, an automated method for generating and curating large-scale binary classification datasets for TinyML, combining multi-source label fusion, confidence-, area-, and depiction-aware filtering, label correction on the evaluation splits, and fine-grained benchmark synthesis.
Applying it to person detection, we released Wake Vision, a dataset of almost 6M images (close to 100$\times$ more person images than \gls{vww}) whose manually relabeled validation and test sets achieve a 2.2\% label error rate, compared to 7.8\% reported for VWW.
Wake Vision-trained models improve test accuracy by up to 6.6\% over \gls{vww} across MobileNetV2, MCUNet, MicroNets, and ColabNAS architectures, match or exceed \gls{vww}-trained models on 13 of 16 fine-grained subsets, and retain their advantage on three out-of-distribution datasets covering driving and overhead-surveillance imagery.

Two findings shape data-centric TinyML beyond person detection. 
First, small models are disproportionately sensitive to training-set label noise. The smallest model loses 1.3\% accuracy when training error rises from 6.8\% to 15\%, versus 0.5\% for the largest. 
Second, two-stage training, i.e., pretraining on the larger noisier set and fine-tuning on the smaller cleaner one, reaches 85.72\% test accuracy versus 84.89\% and 80.80\% for the constituent sets alone.

Looking ahead, we see Wake Vision opening three lines of work for data-centric TinyML.
The first is the long tail of bespoke TinyML tasks. Most TinyML deployments target binary or low-class-count problems for which no public dataset exists, and the Wake Vision pipeline reaches any of the 9.6K trainable classes of Open Images v7 (\Cref{sec:other_binary_datasets}). Bird and car detection are first demonstrations; we expect the pipeline to remove the dataset bottleneck for many more such tasks and to make per-task dataset generation a routine step in the TinyML workflow.
The second is treating the dataset itself as a maintained artifact. The first Wake Vision Challenge already reduced the Wake Vision (Large) label error rate from 15.2\% to 9.8\% in a single round, at a fraction of the \$600,000 implied by manual relabeling at this scale. Subsequent rounds, hosted with the Edge AI Foundation, are intended to keep improving label quality and to fold winning contributions into successive Wake Vision releases, so that progress on the dataset accumulates rather than restarting with each paper.
The third is a research agenda the artifacts make tractable. The label-noise sensitivity above means dataset quality cannot be treated as a solved problem just because larger models tolerate it; tiny models need their own studies. A 6M-image dataset with manually relabeled evaluation splits, fine-grained subsets along demographic and environmental axes, and out-of-distribution probes provides a substrate on which the community can study label-quality-vs-scale trade-offs, robustness, and fairness for tiny models in a controlled way. We release the dataset, benchmarks, code, and the pipeline that produced them under CC-BY 4.0 (\url{https://wakevision.ai}), and invite the TinyML community to build on Wake Vision and to extend the data-centric TinyML toolkit to the tasks beyond person detection where it is most needed.

%% file: sec/appendix.tex
\appendix

\section{Additional Benchmark Results}\label{app:additional_benchmark_results}
Table \ref{tab:additional_benchmark_results} provides input shape, Flash, RAM, and MACs for each model presented in section \ref{sec:comparison_to_vww} as well as the mean and standard deviation of test accuracies for both Wake Vision and \gls{vww} datasets. RAM and Flash have been measured using the ``stm32tflm'' utility of X-CUBE-AI 8.1.0, whereas MACs with ``stm32ai''. 
Links pointing to the original models used for training are also present for each model family involved in the experiment. 

\begin{figure*}[!t]
    \includegraphics[width=\linewidth]{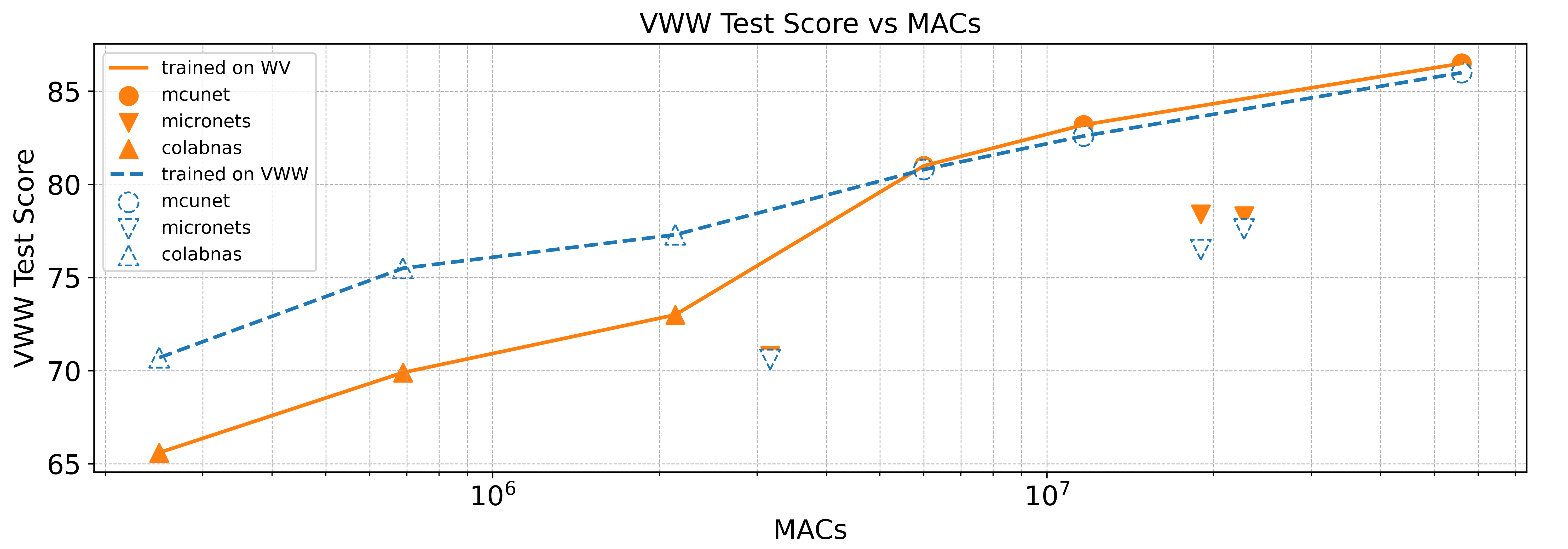}
    \caption{Cross-evaluation results on the Visual Wake Words test set}
    \label{fig:cross-eval-vww}
\end{figure*}

Figure \ref{fig:cross-eval-vww} shows cross-evaluation results on the Visual Wake Words test set. 
The majority of models trained on the Wake Vision dataset obtain better test accuracy on \gls{vww} than the ones trained on \gls{vww}.
Only the smallest models do not achieve a higher test accuracy when trained on Wake Vision, likely due to a lack of learning capacity to overcome the slight domain shift between the two datasets. 

\noindent\begin{table*}[!t]
    \centering
    \caption{Raw data of reconstructed models.}
    \label{tab:additional_benchmark_results}
    \begin{tabular}{c c c c c c c} 
        \toprule
        \bf{Model}                            &  \bf{RAM}                 & \bf{Flash}              & \bf{MACs}                    & \multirow{2}{*}{\bf{Test Set}} & \multicolumn{2}{c}{\bf{Training Set}}  \tabularnewline 
        \bf{.tflite}                          & \bf{[kiB]}               & \bf{[kiB]}              & \bf{[MM]}                   &          & \bf{WV (Qual.)} & \bf{VWW} \tabularnewline \midrule
        \multicolumn{7}{c}{MobileNetV2~\citep{sandler2018mobilenetv2}} \tabularnewline
        \multirow{2}{*}{MobileNetV2\_0.25}    & \multirow{2}{*}{1,244.5} & \multirow{2}{*}{410.55} & \multirow{2}{*}{36,453,732} & \bf{WV}  & $84.9 \pm 0.11$ & $83.8 \pm 0.23$ \tabularnewline
        & & &                                                                                                                                                 & \bf{VWW} & $88.6 \pm 0.17$ & $88.3 \pm 0.29$ \tabularnewline
        \midrule
        \multicolumn{7}{c}{MCUNet~\citep{lin2020mcunet}} \tabularnewline
        \multirow{2}{*}{10fps\_vww}           & \multirow{2}{*}{168.5}   & \multirow{2}{*}{533.84} & \multirow{2}{*}{5,998,334}  & \bf{WV}  & $81.7 \pm 0.28$ & $77.6 \pm 0.31$ \tabularnewline
        & & &                                                                                                                                                 & \bf{VWW} & $81 \pm 0.12$   & $80.8 \pm 0.1$ \tabularnewline
        & & & & & & \tabularnewline
        \multirow{2}{*}{5fps\_vww}            & \multirow{2}{*}{226.5}   & \multirow{2}{*}{624.55} & \multirow{2}{*}{11,645,502} & \bf{WV}  & $82.9 \pm 0.29$ & $79.6 \pm 0.26$ \tabularnewline
        & & &                                                                                                                                                 & \bf{VWW} & $83.2 \pm 0.57$ & $82.6 \pm 0.16$ \tabularnewline
        & & & & & & \tabularnewline
        \multirow{2}{*}{320kb-1mb\_vww}       & \multirow{2}{*}{393}     & \multirow{2}{*}{923.76} & \multirow{2}{*}{56,022,934} & \bf{WV}  & $85.6 \pm 0.34$ & $82.4 \pm 0.62$ \tabularnewline
        & & &                                                                                                                                                 & \bf{VWW} & $86.5 \pm 1.01$ & $86 \pm 0.25$\tabularnewline \midrule
        \multicolumn{7}{c}{MicroNets~\citep{banbury2021micronets}} \tabularnewline
        \multirow{2}{*}{vww2\_50\_50}   & \multirow{2}{*}{71.50}   & \multirow{2}{*}{225.54} & \multirow{2}{*}{3,167,382}  & \bf{WV}  & $71.9 \pm 0.67$ & $66.2 \pm 0.88$ \tabularnewline
        & & &                                                                                                                                                 & \bf{VWW} & $70.8 \pm 0.68$ & $70.6 \pm 0.64$\tabularnewline
        & & & & & & \tabularnewline
        \multirow{2}{*}{vww3\_128\_128} & \multirow{2}{*}{137.50}  & \multirow{2}{*}{463.73} & \multirow{2}{*}{22,690,291} & \bf{WV}  & $77.8 \pm 0.56$ & $72.6 \pm 1.01$ \tabularnewline
        & & &                                                                                                                                                 & \bf{VWW} & $78.3 \pm 0.91$ & $77.6 \pm 1.13$ \tabularnewline
        & & & & & & \tabularnewline
        \multirow{2}{*}{vww4\_128\_128} & \multirow{2}{*}{123.50}  & \multirow{2}{*}{417.03} & \multirow{2}{*}{18,963,302} & \bf{WV}  & $77.9 \pm 0.6$  & $71.3 \pm 1.03$ \tabularnewline
        & & &                                                                                                                                                 & \bf{VWW} & $78.4 \pm 0.51$ & $76.5 \pm 0.37$ \tabularnewline \midrule
        \multicolumn{7}{c}{ColabNAS~\citep{garavagno2024colabnas}} \tabularnewline
        \multirow{2}{*}{k\_2\_c\_3}           & \multirow{2}{*}{18.5}    & \multirow{2}{*}{7.66}   & \multirow{2}{*}{250,256}    & \bf{WV}  & $70.6 \pm 0.96$ & $69.3 \pm 0.97$ \tabularnewline
        & & &                                                                                                                                                 & \bf{VWW} & $65.6 \pm 0.66$ & $70.7 \pm 0.08$ \tabularnewline
        & & & & & & \tabularnewline
        \multirow{2}{*}{k\_4\_c\_5}           & \multirow{2}{*}{22}      & \multirow{2}{*}{18.49}  & \multirow{2}{*}{688,790}    & \bf{WV}  & $75.7 \pm 0.18$ & $74 \pm 0.23$ \tabularnewline
        & & &                                                                                                                                                 & \bf{VWW} & $69.9 \pm 0.26$ & $75.5 \pm 0,64$ \tabularnewline
        & & & & & & \tabularnewline
        \multirow{2}{*}{k\_8\_c\_5}           & \multirow{2}{*}{32.5}    & \multirow{2}{*}{44.56}  & \multirow{2}{*}{2,135,476}  & \bf{WV}  & $77.3 \pm 0.37$ & $75 \pm 0.15$ \tabularnewline
        & & &                                                                                                                                                 & \bf{VWW} & $73 \pm 0.91$   & $77.3 \pm 0.57$ \tabularnewline \bottomrule
        \end{tabular}
\end{table*}

\section{Code Repository}\label{app:code_repository}
The code used to generate the Wake Vision dataset is available at the following GitHub repository: \url{https://github.com/harvard-edge/Wake_Vision}

This repo contains the code to generate Wake Vision and the benchmark suite, as well as the code to train and evaluate models. This code is sufficient to reproduce all results in the paper.

\section{Inducing Label Errors for Quality \& Size Evaluation}
\label{app:label_error_proof}
The goal is to make a single pass through the dataset and flip the labels of a binary classification dataset such that the expected label error rate is $d$.
There is an underlying rate of label errors $e$. If we flip one of these underlying errors, we correct it, thereby inadvertently decreasing the label error rate.
After flipping labels with a probability of $p$ we can claim that the likelihood of a single label being correct is the probability that we flipped the label and it was originally an error plus the probability that we didn't flip the label and it was originally correct: $1-d = p*e + (1-p)(1-e)$.
Then solving for $p$ gives $p = (e-d)/(2e-1)$.
A current flaw of this method is that the injected label errors are not consistent between epochs, which would likely be less destructive to a model's accuracy since the same errors are not reinforced each epoch. This could also potentially explain why large models in Figure \ref{fig:error_vs_dataset_size} don't overfit on training data with higher label errors, as the inconsistent label noise has a regularizing effect.

\begin{figure*}[t]
\centering
    \includegraphics[width=\linewidth]{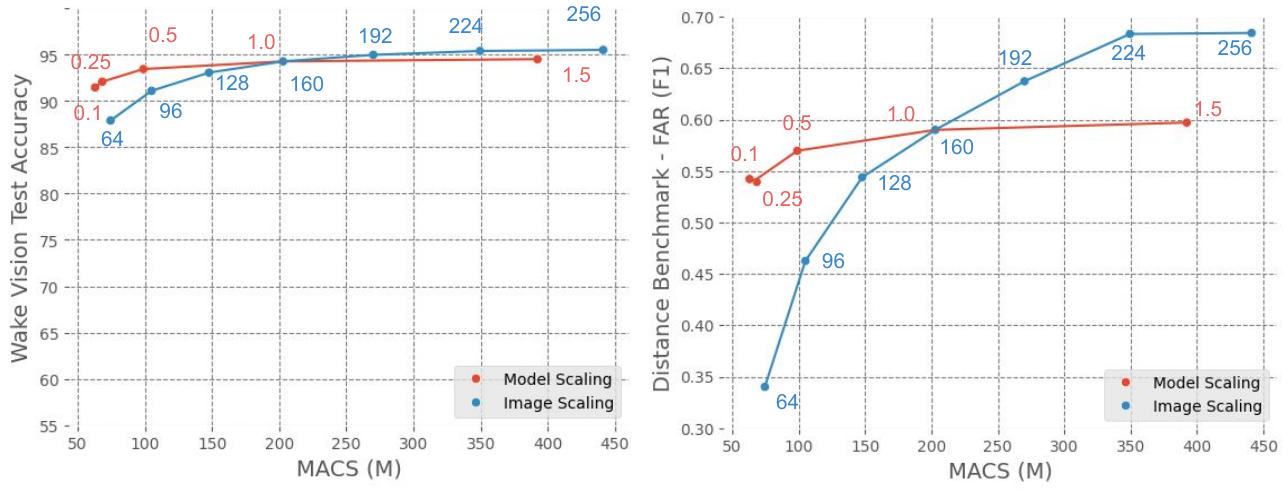}
    \caption{Effects of scaling the image size vs. the model width on Wake Vision test accuracy and the Distance-Far benchmark set.
    The Distance-Far F1 score (right) is far more sensitive to changes in the image size than the overall test accuracy (left).
    Image sizes refer to square image resolutions (e.g. 96x96) and model size refers to the MobileNetV2 width multiplier. When not specified the image size is 160 and the width multiplier is 1.0.}
    \label{fig:scaling-experiment}
\end{figure*}

\section{Model Design Case Study}\label{app:case_study}
Basic test set performance can often misrepresent a model's performance, given the typical domain shift between images scraped from the internet and real-world use cases.
This issue is exacerbated when ML practitioners must trade off accuracy for model performance and size, which is necessary for TinyML use cases.
A design decision might have seemingly little impact on the test accuracy but may destroy real-world performance depending on the deployment environment.
For example, a person detection system may only operate in dark lighting conditions, but the test dataset has an insignificant number of dark samples; therefore, the test accuracy will not reflect the real-world accuracy.

The benchmark suite enables more holistic analysis during the design phase. To show this use we perform a series of scaling experiments employing typical TinyML compression techniques and identifying under which circumstances these techniques are appropriate.
While these results can inform ML practitioners, our intention is to demonstrate the usefulness of the benchmark suite.

\subsection{Image Size vs. Model Width Scaling}
We train two series of MobileNetV2 models: one series that sweeps the input image size [64-256] and one that sweeps the width multiplier of a model [0.1-1.5]. We then benchmark these models on the Wake Vision test set as well as the far distance benchmark. We plot these results against the number of multiply-accumulate (MAC) operations in the model as a proxy for on-device latency~\citep{banbury2021micronets}. 

The results in Figure~\ref{fig:scaling-experiment} (left) show that when looking exclusively at the high-level metric (i.e., test accuracy), scaling the input image size has a similar impact as scaling the model size. However, as shown in Figure~\ref{fig:scaling-experiment} (right), when we consider only samples where the person is far away from the camera (i.e., the distance benchmark), we observe a much more significant impact when scaling the image size. In the case of distant subjects, the image size becomes the bottleneck. 

These findings suggest that for ML developers targeting use cases where the subject is likely to be far from the camera, prioritizing larger input image sizes over wider models may be more beneficial. However, this critical design consideration could be obscured when solely relying on high-level metrics like overall test accuracy. The distance benchmark in Wake Vision effectively unveils the disproportionate impact of image size on model performance for distant subjects, enabling more informed decision-making during model optimization.

\subsection{Quantization}

Quantization is a crucial technique for deploying efficient TinyML models, offering substantial benefits in terms of reduced latency, memory footprint, and model size. However, prior work has suggested that quantization can disproportionately impact the performance of models on underrepresented subsets of data~\citep{hooker2020characterising}. To assess the implications of quantization in the context of person detection, we investigate the impacts of int8 quantization on a model's benchmark results across Wake Vision's fine-grained benchmarks.

Our findings show negligible degradation in performance across all benchmarks ($\pm0.004$ F1) when employing int8 quantization, even on outlier sets. This result contradicts the previously observed disproportionate impact of quantization on underrepresented subsets. We speculate that person detection may be a relatively simple task, potentially explaining why we do not observe this specific property of quantization in our experiments. Given the negligible performance degradation and the substantial latency, memory, and model size benefits of quantization, we conclude that quantization is well-suited for person detection.

\subsection{Grayscale}
\begin{figure}[t!]
\centering
    \includegraphics[width=\linewidth]{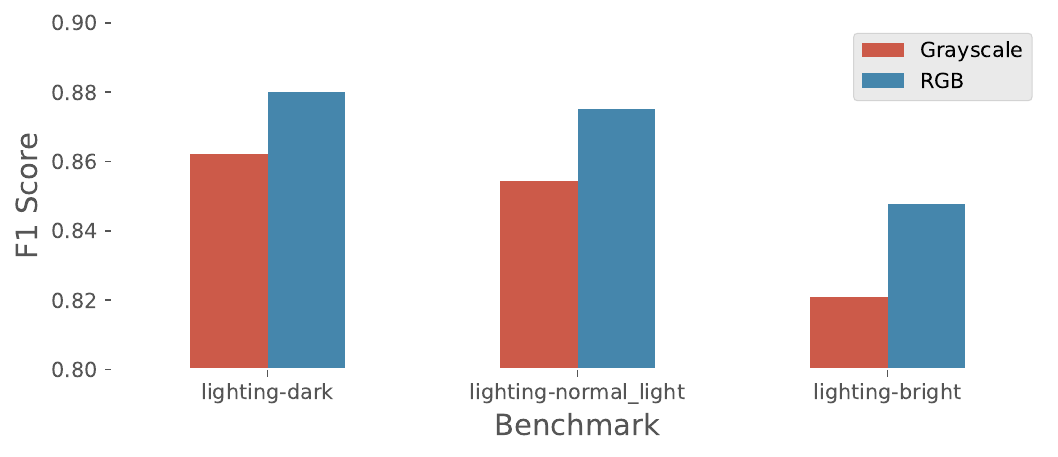}
    \caption{Impact of grayscale input images on the lighting benchmarks: dark, normal light, and bright. Models that use grayscale input images are more sensitive to bright lighting conditions than RGB.}
    \label{fig:grayscale-experiment}
\end{figure}

Converting a model's input image channels to grayscale from RGB is a commonly employed optimization in the TinyML field~\citep{banbury2021micronets} as it can substantially reduce a model's memory consumption.
We observed, however, that the grayscale optimization disproportionately impacts images on the brighter end of the spectrum as illustrated in Figure~\ref{fig:grayscale-experiment}.
This further demonstrates the importance of fine-grained analysis, as some real-world deployment environments might be far brighter than the average Wake Vision test sample.

\section{Open Image Label Distribution}
Wake Vision is derived from Open Images V7, therefore the diversity of subjects in the images should follow the distribution of labels in Open Images. 
The Label distribution of Open images can be found here: \url{https://storage.googleapis.com/openimages/web/factsfigures_v7.html#statistics}

\section{Dataset Access and Organization}\label{app:dataset_access}
Wake Vision is available online at the Harvard Dataverse, via \gls{tfds} and Hugging Face Datasets.
The version on Harvard Dataverse has been assigned the following DOI: \url{https://doi.org/10.7910/DVN/1HOPXC}.
More information can be found on the webpage \url{https://wakevision.ai}.

The dataset is organized as a set of compressed tar files containing images and a series of label CSVs.
The label CSVs are organized such that the file name of the image is the identifying index. 
CSVs for all dataset splits include the person and depiction label.
The Validation and Test label CSVs also have flags that denote a sample's inclusion into a fine-grain benchmark set (e.g. Distance-Near).
This structure makes it easy to access just the required data without requiring a full download.
It also ensures the dataset can be easily updated as new versions are introduced.

\section{Person Label Generation Details}\label{app:label_generation_details}
\begin{description}
    \item{\textbf{Person Labels.}} The most straightforward Open Images label classes to label as a person in Wake Vision are the "person" label and its subcategories (\rv{Configurable, but defaults} listed in Appendix~\ref{sec:person_label_classes}). 
    All of these are relabelled as persons in Wake Vision.
    These labels are present as both image-level labels and bounding box labels.
    
    We furthermore inspect the image-level label classes for synonyms and umbrella terms for all the person related labels.
    This search resulted in an additional six person related label classes.
    These person related label classes will only be used in the image level label configuration.

    \rv{Lastly, the image level person labels in Open Images have an associated confidence score.
    This score ranges from 0-10.
    Labels that have been verified by humans to be absent from an image have a score of 0, while labels that have been verified present by humans have a score of 10.
    Purely machine-generated labels have a fractional confidence score that is generally $>=5$ \citep{OpenImages, OpenImages2}.
    By default, we only respect labels that have a minimum confidence of  7.
    Labels below this threshold are ignored.}
    
    \item{\textbf{Person Body Part Labels.}} Body parts are more challenging to relabel, as it is dependent on the use case whether a body part should be considered a person.
    
    For example a camera that detects whether a person is inside a room to decide if the light should be switched on would want to consider body parts as a person, as this will keep the lights on even when the person is only partly in the camera frame. 
    For waking up electronics, however, it may not make sense to consider body parts as persons. This could, e.g., mean that a computer would turn on when detecting a foot.

    To cater to both use cases we include a flag in our open-source dataset creation code to set whether body parts should be considered persons.
    By default we consider body parts as persons.
    
    \item{\textbf{Depictions.}} Open Images bounding box labels contain metadata about whether an object is a depiction, e.g., a painting or a photograph of a person.
    This presents the challenge of how to handle depictions. 
    While most use cases would not consider a depiction a person, it could make sense to either exclude them to make training easier, or include them as non-persons to make a model resistant to seeing depictions when deployed.

    In line with how we handle body parts, we therefore include a flag for our open-source dataset creation code to set whether depictions should be excluded or considered non-persons.
    By default depictions are considered non-persons.
    
    \item{\textbf{Bounding Box Size.}} The \gls{vww} dataset only considered a \gls{coco} person to be a person if the bounding box around the person took up at least \rv{0.5\%} of the image \citep{chowdhery2019visual}.
    If the person took up less than \rv{0.5\%} of the image, the image was excluded from the dataset.
    \rv{For Wake Vision, we observed a decrease in performance at the 0.5\% limit.
    We theorize that the cause of this decrease is that a person becomes indistinguishable at such a low percentage of already small images.
    Therefore we adopted a 5\% threshold for the Wake Vision dataset instead.}
    For different requirements, users can change a configuration parameter in our open-source dataset creation code.
\end{description}




\section{Open Images Download}\label{sec:open_images_download}
The full Open Images v7 dataset is not hosted by the dataset authors.
Rather it is provided as a collection of Flickr image \glspl{url} and their associated labels.
As an unfortunate result of this, the dataset is not static over time as image owners can delete their images from the Flickr platform.
As a result we were only able to download a subset of the original Open Images v7 dataset as shown in \cref{tab:open_images_download}.

\begin{table}
    \centering
    \caption{Number of images downloaded from Open Images v7. Download occurred between the 28\textsuperscript{th} of November to the 5\textsuperscript{th} of December 2023}
    \label{tab:open_images_download}
    \begin{tabular}{l c c c}
        \toprule
                   & Train     & Validation & Test    \\
        \midrule
        Downloaded & 7,936,979 & 36,406     & 109,305 \\
        Errors     & 1,055,669 & 5,214      & 16,131  \\
        \bottomrule
    \end{tabular}
\end{table}

\section{Person Label Classes}\label{sec:person_label_classes}
\rv{In our default configuration, we} consider the following Open Images v7 labels to be a person for the Wake Vision dataset:
\begin{itemize}[nosep]
    \item \emph{Person}
    \item \emph{Woman} (Subcategory of \emph{Person})
    \item \emph{Man} (Subcategory of \emph{Person})
    \item \emph{Girl} (Subcategory of \emph{Person})
    \item \emph{Boy} (Subcategory of \emph{Person})
    \item \emph{Human body} (Part of \emph{Person})
    \item \emph{Human face} (Part of \emph{Person})
    \item \emph{Human head} (Part of \emph{Person})
    \item \emph{Human} (Person synonym - Only in Image Level Label Configuration)
    \item \emph{Female person} (Woman synonym - Only in Image Level Label Configuration)
    \item \emph{Male person} (Man synonym - Only in Image Level Label Configuration)
    \item \emph{Child} (Umbrella term for \emph{Girl} \& \emph{Boy} - Only in Image Level Label Configuration)
    \item \emph{Adolescent} (Umbrella term for \emph{Girl} \& \emph{Boy} - Only in Image Level Label Configuration)
    \item \emph{Youth} (Umbrella term for \emph{Girl} \& \emph{Boy} - Only in Image Level Label Configuration)
\end{itemize}

\rv{The following labels are considered body parts and are labeled according to the body part configuration}:
\begin{itemize}[nosep]
    \item \emph{Human eye} (Part of \emph{Person})
    \item \emph{Skull} (Part of \emph{Person})
    \item \emph{Human mouth} (Part of \emph{Person})
    \item \emph{Human ear} (Part of \emph{Person})
    \item \emph{Human nose} (Part of \emph{Person})
    \item \emph{Human hair} (Part of \emph{Person})
    \item \emph{Human hand} (Part of \emph{Person})
    \item \emph{Human foot} (Part of \emph{Person})
    \item \emph{Human arm} (Part of \emph{Person})
    \item \emph{Human leg} (Part of \emph{Person})
    \item \emph{Beard} (Part of \emph{Person})
\end{itemize}


\section{Manual Label Correction}\label{app:manual-label-correction}
\begin{figure*}
    \centering
    \includegraphics[width=\linewidth]{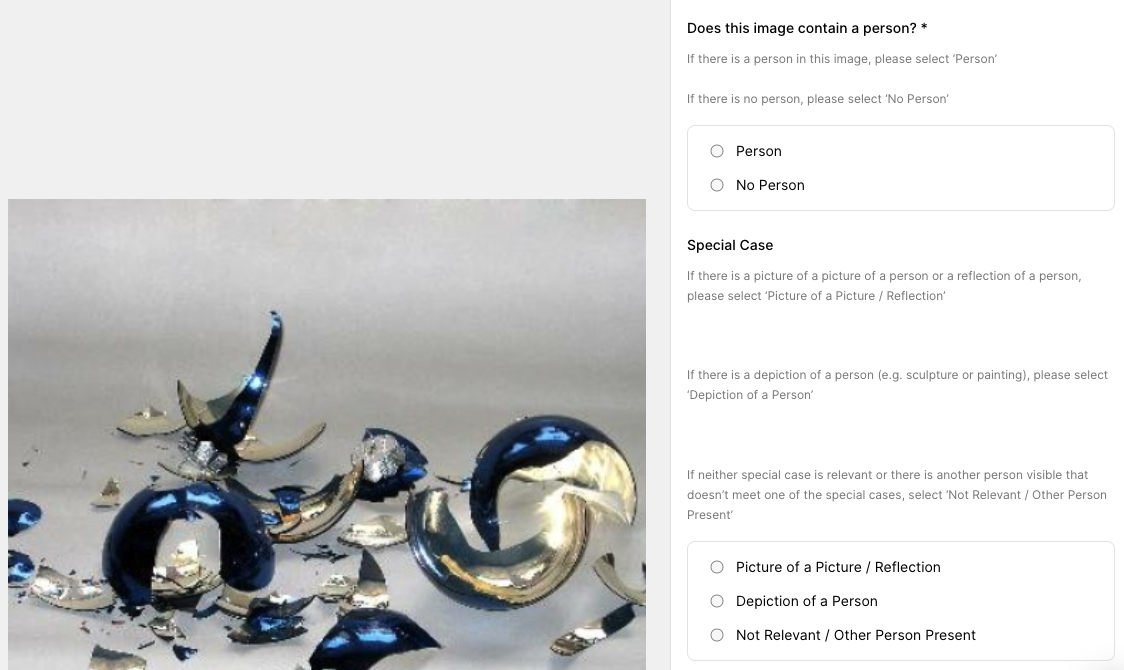}
    \caption{A screenshot of the labeling menu used to manually label the validation and test sets.}
    \label{fig:labeling-screenshot}
\end{figure*}

We use the Scale Rapid tool from Scale AI to relabel the Wake Vision validation and test sets.
Figure \ref{fig:labeling-screenshot} shows a screenshot of the labeling menu presented to the human labelers.
The labelers were instructed to label an image as a "person" if a person was present anywhere in the image and "no person" if no visible person was present.
The labelers also indicated if the image contained a depiction of a person or a "picture of a picture/reflection", which we use for metadata.

The cost per image was \$0.10 USD and is set based on Scale's pricing structure. 
Each image is labeled by 3 different labelers to form a consensus.
The total cost of labeling was \$7089.8 USD.
The authors do not know the hourly rate paid to each labeler, but Scale lists \$18 USD/hr on job postings at the time of writing. 
For context on the difficulty of the task, the authors were able to average around 500 images per hour at a reasonable pace when estimating error rates.

\section{Automatic Label Correction}
\label{app:auto-label-correction}
To address the challenge of correcting Wake Vision validation and test labels we initially employed the Confident Learning technique~\citep{northcutt2021confident} to intelligently identify potential label errors.
We selected Confident Learning as prior work has demonstrated its capability to find label errors in large datasets~\citep{northcutt2021pervasive}. 
Confident Learning flagged suspected mislabeled instances, which we inspected and corrected through a manual verification process.

\begin{table}[t]
    \centering
    \caption{Amount of label errors identified and corrected using Confident Learning.}
    \label{tab:confident learning}
    \setlength{\tabcolsep}{4pt}
    \begin{tabular}{lcccc}
        \toprule
        Dataset & Total & Suggested & Corrected & Suggestion\\
        Split & Size & Errors & Errors & Accuracy\\
        \midrule
        Validation & 18,582 & 632 & 81 & 12.82\%\\
        Test & 51,282 & 1672 & 267 & 15.97\%\\
        \bottomrule
    \end{tabular}
\end{table}

As shown in Table~\ref{tab:confident learning}, the confident learning process identified a large amount of possible label errors in Wake Vision's Validation and Test sets.
However, only between 12 and 16\% of these possible label errors were legitimate errors.
We corrected these identified label errors in the Wake Vision validation and test sets.

Given this low acceptance rate of label issues identified by Confident Learning, we concluded that we could not automate label cleaning to a point where no human in the loop is needed.
This made this strategy too human-intensive to be applied to the much larger training sets.

To further correct label errors in the final Wake Vision Validation and Test sets we employed the Scale AI platform to crowd-source manual label corrections. 
These label corrections are described in \cref{sec:label_correction} and deprecate the automatic label corrections described in this section.

\section{Author Statement}
The authors of this paper hereby confirm that we bear all responsibility in case of violation of rights, etc, and confirm the data license to be CC BY 4.0 for our labels, and that all images in the dataset have been uploaded to Flickr with a CC BY 2.0 license.
Note: while the Open Images authors tried to identify images that are licensed under a Creative Commons Attribution license, we make no representations or warranties regarding the license status of each image and users should verify the license for each image themselves.

\section{Hosting, License and Maintenance Plan}
The dataset is hosted on Harvard Dataverse and HuggingFace Datasets. Additionally, the dataset is available through TensorFlow datasets and can be regenerated from Open Images using our open-source filtering code. 
The Edge AI Foundation will assume responsibility for hosting and maintaining the dataset long-term.
In Appendix~\ref{app:image-removal} we describe our procedure for image take-downs if individuals wish to remove their likeness from the dataset.

Wake Vision's labels and metadata are licensed under a CC BY 4.0 license. 
Images in the dataset have all been uploaded to Flickr with a CC BY 2.0 license.
Note: while the Open Images authors tried to identify images that are licensed under a Creative Commons Attribution license, we make no representations or warranties regarding the license status of each image and users should verify the license for each image themselves.

\section{DOI and Croissant URLs}
The version of Wake Vision on Harvard Dataverse has been assigned the following DOI: \url{https://doi.org/10.7910/DVN/1HOPXC}.
The following two croissant files describe the Wake Vision (Quality, Validation, and Test) and Wake Vision (Large) respectively:
\url{https://huggingface.co/api/datasets/Harvard-Edge/Wake-Vision/croissant},

\noindent\sloppy
\url{https://huggingface.co/api/datasets/Harvard-Edge/Wake-Vision-Train-Large/croissant}

\new{\section{Wake Vision Challenge Results}\label{app:data_centric_competition_experiments}}
Submissions to the model-centric track have been trained using the same recipe adopted in \cref{sec:comprehensive_model_architecture_evaluation}. Table \ref{tab:model_centric_track_results} provides input shape, Flash, RAM, and MACs for each model presented in \cref{sec:wake_vision_challenge} as well as the mean and standard deviation of test accuracies for the Wake Vision dataset. RAM and Flash have been measured using the ``stm32tflm'' utility of X-CUBE-AI 8.1.0, whereas MACs with ``stm32ai''. 

\noindent\begin{table}[h]    
    \centering
    \caption{Raw data of model-centric track.}
    \label{tab:model_centric_track_results}
    \begin{tabular}{c c c c c c} 
        \toprule
        \bf{Model}           & \bf{input}  & \bf{RAM}   & \bf{Flash} & \bf{MACs}  & \bf{Test Acc}  \tabularnewline 
        \bf{.tflite}         & \bf{volume} & \bf{[kiB]} & \bf{[kiB]} & \bf{[MM]}  & \bf{[\%]}      \tabularnewline \midrule
        anas-benalla         & (50,50,3)   & 48.5       & 104.55     & 4,899,292  &  $79.7 \pm 0.28$ \tabularnewline
        samy                 & (80,80,3)   & 73.5       & 34.55      & 5,718,064  &  $79.5 \pm 0.61$ \tabularnewline
        benx13               & (48,48,3)   & 63.5       & 94.95      & 1,693,474  &  $79.6 \pm 0.22$ \tabularnewline
        cezar                & (50,50,3)   & 245.5      & 57.23      & 20,435,868 &  $79.6 \pm 0.42$ \tabularnewline
        mohammad\_hallaq     & (80,80,3)   & 129.5      & 128.32     & 3,887,331  &  $77.3 \pm 0.5$  \tabularnewline
        ymac                 & (50,50,3)   & 34.5       & 27.77      & 431,985    &  $76.7 \pm 0.51$ \tabularnewline
        apighetti            & (50,50,3)   & 48.5       & 278.36     & 693,818    &  $74.3 \pm 0.31$ \tabularnewline
        \bottomrule
        \end{tabular}
\end{table}

Table \ref{tab:data_centric_track_results} shows test accuracy, precision, and recall of a MobileNetV2 having a width modifier of 0.25, trained on the original Wake Vision large split and on the same split enhanced with the technique proposed by the winner of the data-centric track~\citep{WakeVisionChallenge2025_Bedioui}. The training recipe is the same reported in \cref{sec:comprehensive_model_architecture_evaluation}.

\noindent\begin{table}[h]    
    \centering
    \caption{Main improvement of the data-centric track.}
    \label{tab:data_centric_track_results}
    \begin{tabular}{c c c} 
        \toprule
        \multirow{2}{*}{\bf{Metrics}} &  \multicolumn{2}{c}{\bf{Train Large Split}} \tabularnewline 
                               & \bf{Original}   & \bf{Enhanced}             \tabularnewline \midrule
        \multicolumn{3}{c}{MobileNetV2\_0.25}                                 \tabularnewline                       
        Test Accuracy          & $82.3 \pm 0.2$  & \bf{$87.1 \pm 0.81$}      \tabularnewline
        Precision              & $86.4 \pm 1.49$ & \bf{$87.6 \pm 1.78$}      \tabularnewline
        Recall                 & $76.7 \pm 1.83$ & \bf{$86.5 \pm 1.32$}      \tabularnewline
        \bottomrule
        \end{tabular}
\end{table}

\section{Removal of Images}\label{app:image-removal}
Removal of images can be asked via email at \href{mailto:wakevision@edgeaifoundation.org}{wakevision@edgeaifoundation.org}.

%% file: sec/datasheet.tex
\newcommand{\sectioncolor}{black}

\section{Datasheet for Wake Vision}

\noindent
\textbf{
This document is based on \textit{Datasheets for Datasets} by Gebru \textit{et
al.} \citep{gebru2021datasheets}. Please see the most updated version
\underline{\textcolor{blue}{\href{http://arxiv.org/abs/1803.09010}{here}}}.
}

\begin{mdframed}[linecolor=\sectioncolor]
\section*{\textcolor{\sectioncolor}{
    MOTIVATION
}}
\end{mdframed}

    \textcolor{\sectioncolor}{\textbf{
    For what purpose was the dataset created?
    }
    Was there a specific task in mind? Was there
    a specific gap that needed to be filled? Please provide a description.
    } \\
    
    Wake Vision was created as a prototype for a new generation of datasets for TinyML. 
    Where previous TinyML datasets were small due to limited resources, Wake Vision is a production-grade dataset consisting of almost 6 million images and with an extensively cleaned validation and test set.
    Wake Vision specifically targets the binary person classification task.
    Such a large dataset enables TinyML researchers to explore questions previously impossible.
    For example, the size of the Wake Vision dataset enables new research into the tradeoff that often exists between dataset size and sample quality.\\
    
    \textcolor{\sectioncolor}{\textbf{
    Who created this dataset (e.g., which team, research group) and on behalf
    of which entity (e.g., company, institution, organization)?
    }
    } \\
    
    The authors of this paper created the dataset. 
    \\
    
    \textcolor{\sectioncolor}{\textbf{
    What support was needed to make this dataset?
    }
    (e.g. who funded the creation of the dataset? If there is an associated
    grant, provide the name of the grantor and the grant name and number, or if
    it was supported by a company or government agency, give those details.)
    } \\
    
    The Google TPU Research Cloud program partially supported this work via cloud computing credits. 
    The PhD stipend of Emil Njor was supported by the Innovation Fund Denmark DIREC project (9142-00001B).
    Emil Njor's external research stay was furthermore supported by Fulbright Denmark, Stibo Fonden, Thomas. B. Thriges Fond, Otto Mønsteds Fond and Kaj og Hermilla Ostenfelds Fond.
    This work was also supported by the National Science Foundation (NSF) and the Semiconductor Research Corporation (SRC).\\
    
    \textcolor{\sectioncolor}{\textbf{
    Any other comments?
    }} \\
    No additional comments \\

\begin{mdframed}[linecolor=\sectioncolor]
\section*{\textcolor{\sectioncolor}{
    COMPOSITION
}}
\end{mdframed}
    \textcolor{\sectioncolor}{\textbf{
    What do the instances that comprise the dataset represent (e.g., documents,
    photos, people, countries)?
    }
    Are there multiple types of instances (e.g., movies, users, and ratings;
    people and interactions between them; nodes and edges)? Please provide a
    description.
    } \\
    
    The dataset is composed of images, labels, and metadata. The images are sourced from Open Images. The labels are binary, with `1' indicating there is a person somewhere in the image and `0' indicating that there is not a person in the image. The metadata is used to form the benchmark suite where we can evaluate models in specific challenging settings.
    Each image has a binary variable for each fine-grained benchmark where a `1' indicates that the image belongs to the fine-grained benchmark set.\\
    
    \textcolor{\sectioncolor}{\textbf{
    How many instances are there in total (of each type, if appropriate)?
    }
    } \\
   
    There are 5,760,428 labeled images in the dataset. \\
    
    \textcolor{\sectioncolor}{\textbf{
    Does the dataset contain all possible instances or is it a sample (not
    necessarily random) of instances from a larger set?
    }
    If the dataset is a sample, then what is the larger set? Is the sample
    representative of the larger set (e.g., geographic coverage)? If so, please
    describe how this representativeness was validated/verified. If it is not
    representative of the larger set, please describe why not (e.g., to cover a
    more diverse range of instances, because instances were withheld or
    unavailable).
    } \\

    The dataset uses a large subset of the Open Images dataset (6 million out of a possible 9 million images). We use a subset in order to have a balanced number of person samples and non-person samples. We also filter the dataset for quality. The filtering removes subjects that are too far and outside of a center crop. Open Images sources images from Flickr that are licensed as CC-BY 2.0. This is likely not a fully representative sample of the overall world population and likely skews heavily towards a European or North American population. See the Open Images paper~\citep{OpenImages} for more information. \\
    
    \textcolor{\sectioncolor}{\textbf{
    What data does each instance consist of?
    }
    “Raw” data (e.g., unprocessed text or images) or features? In either case,
    please provide a description.
    } \\
    
    The dataset is composed of images, labels, and metadata. \\
    
    \textcolor{\sectioncolor}{\textbf{
    Is there a label or target associated with each instance?
    }
    If so, please provide a description.
    } \\
    
    Yes, every image has a binary label to indicate if a person is visible in the image.
    Images in some dataset splits include more metadata labels, e.g., about the distance of the subject or whether the subject is a depiction.\\
    
    \textcolor{\sectioncolor}{\textbf{
    Is any information missing from individual instances?
    }
    If so, please provide a description, explaining why this information is
    missing (e.g., because it was unavailable). This does not include
    intentionally removed information, but might include, e.g., redacted text.
    } \\

    The training set does not contain the metadata used for the fine-grained benchmark suite.
    The dataset contains two training sets, one of higher quality and one larger.
    The higher quality dataset and other sets contain information about depictions and body parts. 
    The large training set does not contain this information.\\
    
    \textcolor{\sectioncolor}{\textbf{
    Are relationships between individual instances made explicit (e.g., users’
    movie ratings, social network links)?
    }
    If so, please describe how these relationships are made explicit.
    } \\
    
    No \\
    
    \textcolor{\sectioncolor}{\textbf{
    Are there recommended data splits (e.g., training, development/validation,
    testing)?
    }
    If so, please provide a description of these splits, explaining the
    rationale behind them.
    } \\
    
    Yes there are separate splits for training, validation, and testing. These splits are generated from equivalent splits in Open Images. 
    The dataset contains two training splits, one focusing on quality of samples and one on number of samples.
    \\
    
    \textcolor{\sectioncolor}{\textbf{
    Are there any errors, sources of noise, or redundancies in the dataset?
    }
    If so, please provide a description.
    } \\
    
    Yes there are label errors in the dataset due to the labeling process. We discuss this at length in the paper. \\
    
    \textcolor{\sectioncolor}{\textbf{
    Is the dataset self-contained, or does it link to or otherwise rely on
    external resources (e.g., websites, tweets, other datasets)?
    }
    If it links to or relies on external resources, a) are there guarantees
    that they will exist, and remain constant, over time; b) are there official
    archival versions of the complete dataset (i.e., including the external
    resources as they existed at the time the dataset was created); c) are
    there any restrictions (e.g., licenses, fees) associated with any of the
    external resources that might apply to a future user? Please provide
    descriptions of all external resources and any restrictions associated with
    them, as well as links or other access points, as appropriate.
    } \\
    
    The dataset is self-contained. We re-host the images, labels, and metadata.
    The images may be subject to different licenses than our labels and metadata. 
    All images have originally been hosted on Flickr under a permissive CC BY license. \\
    
    \textcolor{\sectioncolor}{\textbf{
    Does the dataset contain data that might be considered confidential (e.g.,
    data that is protected by legal privilege or by doctor-patient
    confidentiality, data that includes the content of individuals’ non-public
    communications)?
    }
    If so, please provide a description.
    } \\
    
    No, the images were posted online with the CC-BY 2.0 license. It is possible an image was posted without the individual's consent. While we tried to identify images that are licensed under a Creative Commons Attribution license, we make no representations or warranties regarding the license status of each image, and a user should verify the license for each image themselves. \\
    
    \textcolor{\sectioncolor}{\textbf{
    Does the dataset contain data that, if viewed directly, might be offensive,
    insulting, threatening, or might otherwise cause anxiety?
    }
    If so, please describe why.
    } \\
    
    No, not to our knowledge. Due to the size of the dataset we have not had the resources to manually verify this. \\
    
    \textcolor{\sectioncolor}{\textbf{
    Does the dataset relate to people?
    }
    If not, you may skip the remaining questions in this section.
    } \\
    
    Yes, the dataset contains images of people. \\
    
    \textcolor{\sectioncolor}{\textbf{
    Does the dataset identify any subpopulations (e.g., by age, gender)?
    }
    If so, please describe how these subpopulations are identified and
    provide a description of their respective distributions within the dataset.
    } \\
    
    Yes the benchmark suite contains subsets that focus on perceived age and gender. These labels have been sourced from the More Inclusive Annotations of People~\citep{miap_aies}. The purpose of these benchmarks is to test the fairness of a trained model across these subpopulation and we do not condone using these labels to train a model. The size of these sets is listed in Table \ref{tab:benchmark_suite_transposed_compact}.\\
    
    \textcolor{\sectioncolor}{\textbf{
    Is it possible to identify individuals (i.e., one or more natural persons),
    either directly or indirectly (i.e., in combination with other data) from
    the dataset?
    }
    If so, please describe how.
    } \\
    
    It could be possible to identify a person by their face or other visible characteristics in the image. We do not share any other personal information besides the images. \\
    
    \textcolor{\sectioncolor}{\textbf{
    Does the dataset contain data that might be considered sensitive in any way
    (e.g., data that reveals racial or ethnic origins, sexual orientations,
    religious beliefs, political opinions or union memberships, or locations;
    financial or health data; biometric or genetic data; forms of government
    identification, such as social security numbers; criminal history)?
    }
    If so, please provide a description.
    } \\
    
    No, not to our knowledge. Some personal or sensitive information may be visible in the images. However, the images were shared publicly under a CC-BY 2.0 license. We do not include any personal information beyond the images. \\
    
    \textcolor{\sectioncolor}{\textbf{
    Any other comments?
    }} \\
    
    The images used were published publicly under a CC-BY 2.0 license and are already used in the Open Images~\citep{OpenImages2} dataset and its derivatives. The authors of Open images tried to identify images that are licensed under a Creative Commons Attribution license but make no representations or warranties regarding the license status of each image and a user should verify the license for each image themselves.\\

\begin{mdframed}[linecolor=\sectioncolor]
\section*{\textcolor{\sectioncolor}{
    COLLECTION
}}
\end{mdframed}

    \textcolor{\sectioncolor}{\textbf{
    How was the data associated with each instance acquired?
    }
    Was the data directly observable (e.g., raw text, movie ratings),
    reported by subjects (e.g., survey responses), or indirectly
    inferred/derived from other data (e.g., part-of-speech tags, model-based
    guesses for age or language)? If data was reported by subjects or
    indirectly inferred/derived from other data, was the data
    validated/verified? If so, please describe how.
    } \\
    
    The dataset is derived from Open Images~\citep{OpenImages2}, which in turn acquired its images from Flickr. Section \ref{sec:labeling} describes the process.\\
    
    \textcolor{\sectioncolor}{\textbf{
    Over what timeframe was the data collected?
    }
    Does this timeframe match the creation timeframe of the data associated
    with the instances (e.g., recent crawl of old news articles)? If not,
    please describe the timeframe in which the data associated with the
    instances was created. Finally, list when the dataset was first published.
    } \\
    
    The original data collection was conducted in November of 2015. The Open Images paper~\citep{OpenImages} details the image acquisition process. \\
    
    \textcolor{\sectioncolor}{\textbf{
    What mechanisms or procedures were used to collect the data (e.g., hardware
    apparatus or sensor, manual human curation, software program, software
    API)?
    }
    How were these mechanisms or procedures validated?
    } \\
    
    The images were sources from Flickr in an automated pipeline. The original labeling pipeline was a combination of automated labeling and manual verification and labeling. The Open Images paper details this process~\citep{OpenImages}.
    We detail our process for relabeling and filtering the dataset in Section~\ref{sec:labeling}.\\
    
    \textcolor{\sectioncolor}{\textbf{
    What was the resource cost of collecting the data?
    }
    (e.g. what were the required computational resources, and the associated
    financial costs, and energy consumption - estimate the carbon footprint.
    See Strubell \textit{et al.}~\citep{strubell2019energy} for approaches in this area.)
    } \\
    
    We used cloud TPU credits provided by the Google Cloud TRC program. The primary source of required resources was model training to evaluate the dataset and the storage and bandwidth required to process and upload the dataset to hosting locations.
    The total size of the dataset is approximately 2 TB.
    To the authors' knowledge, there is no public information about the power draw of TPUs. 
    This matches the conclusions drawn by Strubell~\textit{et al.} \citep{strubell2019energy}.
    This makes it impossible to disclose the energy consumption and estimated carbon footprint of the project.
    
    We used paid crowdsource labelers to relabel the validation and test sets. The total cost of labeling was approximately 7,000 USD. \\
    
    \textcolor{\sectioncolor}{\textbf{
    If the dataset is a sample from a larger set, what was the sampling
    strategy (e.g., deterministic, probabilistic with specific sampling
    probabilities)?
    }
    } \\
    
    The dataset is a large subset of the total Open Images dataset. The subset is generated by a deterministic filtering process described in Section \ref{sec:labeling}. \\
    
    \textcolor{\sectioncolor}{\textbf{
    Who was involved in the data collection process (e.g., students,
    crowdworkers, contractors) and how were they compensated (e.g., how much
    were crowdworkers paid)?
    }
    } \\
    
    We used paid crowdsource labelers to relabel the validation and test sets. The total cost of labeling was approximately 7,000 USD.
    We used the Scale Rapid labeling service. A total of 70,898 images were labeled at a cost of 0.1 USD per image and 0.033 USD per individual label.
    Appendix \ref{sec:label_correction} gives additional detail.\\
    
    \textcolor{\sectioncolor}{\textbf{
    Were any ethical review processes conducted (e.g., by an institutional
    review board)?
    }
    If so, please provide a description of these review processes, including
    the outcomes, as well as a link or other access point to any supporting
    documentation.
    } \\
    
    No \\
    
    \textcolor{\sectioncolor}{\textbf{
    Does the dataset relate to people?
    }
    If not, you may skip the remainder of the questions in this section.
    } \\
    
    Yes the dataset contains images of people. \\
    
    \textcolor{\sectioncolor}{\textbf{
    Did you collect the data from the individuals in question directly, or
    obtain it via third parties or other sources (e.g., websites)?
    }
    } \\
    
    We obtained the images via the Open Images dataset~\citep{OpenImages} which originally sourced its images from Flickr~\citep{flickr_webpage}.\\
    
    \textcolor{\sectioncolor}{\textbf{
    Were the individuals in question notified about the data collection?
    }
    If so, please describe (or show with screenshots or other information) how
    notice was provided, and provide a link or other access point to, or
    otherwise reproduce, the exact language of the notification itself.
    } \\
    
    No the individuals were not notified to our knowledge. The images were posted publicly under the CC-BY 2.0 license. \\
    
    \textcolor{\sectioncolor}{\textbf{
    Did the individuals in question consent to the collection and use of their
    data?
    }
    If so, please describe (or show with screenshots or other information) how
    consent was requested and provided, and provide a link or other access
    point to, or otherwise reproduce, the exact language to which the
    individuals consented.
    } \\
    
    Yes, the images were posted publicly under the CC-BY 2.0 license. It is possible that the images were uploaded by a third party without the consent of the individual in the image. \\
    
    \textcolor{\sectioncolor}{\textbf{
    If consent was obtained, were the consenting individuals provided with a
    mechanism to revoke their consent in the future or for certain uses?
    }
     If so, please provide a description, as well as a link or other access
     point to the mechanism (if appropriate)
    } \\
    
    Appendix~\ref{app:image-removal} describes how we can be contacted for removal of individual images. 
    As with Open Images, we make no representations or warranties regarding the license status of each image, and a user should verify the license for each image themselves via the Flickr link in the metadata.
    Note that it is impossible to revoke a CC BY license after publication \citep{cc_by_revoke_information}.\\
    
    \textcolor{\sectioncolor}{\textbf{
    Has an analysis of the potential impact of the dataset and its use on data
    subjects (e.g., a data protection impact analysis)been conducted?
    }
    If so, please provide a description of this analysis, including the
    outcomes, as well as a link or other access point to any supporting
    documentation.
    } \\
    
    No formal analysis has been conducted. We describe the ethical considerations of the dataset in Section \ref{sec:ethical-considerations}.\\
    
    \textcolor{\sectioncolor}{\textbf{
    Any other comments?
    }} \\
    
    No additional comments. \\

\begin{mdframed}[linecolor=\sectioncolor]
\section*{\textcolor{\sectioncolor}{
    PREPROCESSING / CLEANING / LABELING
}}
\end{mdframed}

    \textcolor{\sectioncolor}{\textbf{
    Was any preprocessing/cleaning/labeling of the data
    done(e.g.,discretization or bucketing, tokenization, part-of-speech
    tagging, SIFT feature extraction, removal of instances, processing of
    missing values)?
    }
    If so, please provide a description. If not, you may skip the remainder of
    the questions in this section.
    } \\
    
    Yes we filtered the data as shown in Figure \ref{fig:filter-flowchart}.\\

    \textcolor{\sectioncolor}{\textbf{
    Was the “raw” data saved in addition to the preprocessed/cleaned/labeled
    data (e.g., to support unanticipated future uses)?
    }
    If so, please provide a link or other access point to the “raw” data.
    } \\
    
    The ``raw" data is accessible in the Open Images dataset.\\

    \textcolor{\sectioncolor}{\textbf{
    Is the software used to preprocess/clean/label the instances available?
    }
    If so, please provide a link or other access point.
    } \\
    
    Yes, we open-source our filtering code. \url{https://github.com/harvard-edge/Wake_Vision} \\

    \textcolor{\sectioncolor}{\textbf{
    Any other comments?
    }} \\
    
    No additional comments \\

\begin{mdframed}[linecolor=\sectioncolor]
\section*{\textcolor{\sectioncolor}{
    USES
}}
\end{mdframed}

    \textcolor{\sectioncolor}{\textbf{
    Has the dataset been used for any tasks already?
    }
    If so, please provide a description.
    } \\
    
    The dataset has been used for the experiments reported in the paper. \\

    \textcolor{\sectioncolor}{\textbf{
    Is there a repository that links to any or all papers or systems that use the dataset?
    }
    If so, please provide a link or other access point.
    } \\
    
    The usage of the dataset can be tracked via the citations of this paper. We also host competitions using the dataset and track participants. \\

    \textcolor{\sectioncolor}{\textbf{
    What (other) tasks could the dataset be used for?
    }
    } \\
    
    The dataset can be used as a benchmark for TinyML models and as the basis for data-centric AI research given the prevalence of label errors in the training set and the cleaned nature of the validation and test sets. \\

    \textcolor{\sectioncolor}{\textbf{
    Is there anything about the composition of the dataset or the way it was
    collected and preprocessed/cleaned/labeled that might impact future uses?
    }
    For example, is there anything that a future user might need to know to
    avoid uses that could result in unfair treatment of individuals or groups
    (e.g., stereotyping, quality of service issues) or other undesirable harms
    (e.g., financial harms, legal risks) If so, please provide a description.
    Is there anything a future user could do to mitigate these undesirable
    harms?
    } \\
    
    Yes, the estimated error rates are reported in Table \ref{tab:wv-vww-test-errors}.
    We also discuss the high likelihood of false negatives in Wake Vision's training sets in Section~\ref{sec:label_correction}.
    These errors are due to the labeling and filtering process. We mitigate the impact of these errors by manually labeling the validation and test sets as well as study their impact for TinyML models in Section \ref{sec:quality_vs_size}.\\

    \textcolor{\sectioncolor}{\textbf{
    Are there tasks for which the dataset should not be used?
    }
    If so, please provide a description.
    } \\
    
    The perceived age and gender benchmarks should not be used to train a gender or age classifier. \\

    \textcolor{\sectioncolor}{\textbf{
    Any other comments?
    }} \\
    
    No additional comments \\

\begin{mdframed}[linecolor=\sectioncolor]
\section*{\textcolor{\sectioncolor}{
    DISTRIBUTION
}}
\end{mdframed}

    \textcolor{\sectioncolor}{\textbf{
    Will the dataset be distributed to third parties outside of the entity
    (e.g., company, institution, organization) on behalf of which the dataset
    was created?
    }
    If so, please provide a description.
    } \\
    
    The dataset is hosted on Harvard Dataverse and HuggingFace Datasets. Additionally, the dataset is available through TensorFlow datasets and can be regenerated from Open Images with our open source filtering code.\\

    \textcolor{\sectioncolor}{\textbf{
    How will the dataset will be distributed (e.g., tarball on website, API,
    GitHub)?
    }
    Does the dataset have a digital object identifier (DOI)?
    } \\
    
    Yes, the dataset is hosted on Harvard Dataverse and HuggingFace Datasets. Additionally, the dataset is available through TensorFlow datasets and can be regenerated from Open Images with our open source filtering code. \\

    \textcolor{\sectioncolor}{\textbf{
    When will the dataset be distributed?
    }
    } \\
    
    The Dataset is publicly accessible at the time of writing.\\

    \textcolor{\sectioncolor}{\textbf{
    Will the dataset be distributed under a copyright or other intellectual
    property (IP) license, and/or under applicable terms of use (ToU)?
    }
    If so, please describe this license and/or ToU, and provide a link or other
    access point to, or otherwise reproduce, any relevant licensing terms or
    ToU, as well as any fees associated with these restrictions.
    } \\
    
    The dataset labels are published under the CC-BY 4.0 license. 
    All images in the dataset have been uploaded to Flickr under a CC BY 2.0 license.
    Note: while the Open Images authors tried to identify images that are licensed under a Creative Commons Attribution license, we make no representations or warranties regarding the license status of each image and users should verify the license for each image themselves.\\

    \textcolor{\sectioncolor}{\textbf{
    Have any third parties imposed IP-based or other restrictions on the data
    associated with the instances?
    }
    If so, please describe these restrictions, and provide a link or other
    access point to, or otherwise reproduce, any relevant licensing terms, as
    well as any fees associated with these restrictions.
    } \\
    
    No \\

    \textcolor{\sectioncolor}{\textbf{
    Do any export controls or other regulatory restrictions apply to the
    dataset or to individual instances?
    }
    If so, please describe these restrictions, and provide a link or other
    access point to, or otherwise reproduce, any supporting documentation.
    } \\
    
    No \\

    \textcolor{\sectioncolor}{\textbf{
    Any other comments?
    }} \\
    
    No additional comments. \\

\begin{mdframed}[linecolor=\sectioncolor]
\section*{\textcolor{\sectioncolor}{
    MAINTENANCE
}}
\end{mdframed}

    \textcolor{\sectioncolor}{\textbf{
    Who is supporting/hosting/maintaining the dataset?
    }
    } \\
    
    The dataset is hosted on
    HuggingFace Datasets.
    Additionally, the dataset is available through TensorFlow datasets and can be regenerated from Open Images with our open-source filtering code. 
    The Edge AI Foundation will assume responsibility for hosting and maintaining the dataset long-term. \\

    \textcolor{\sectioncolor}{\textbf{
    How can the owner/curator/manager of the dataset be contacted (e.g., email
    address)?
    }
    } \\
    
    At the time of writing, the authors can be contacted at cbanbury@g.harvard.edu. Once maintenance is transferred to the Edge AI Foundation, there will be a point of contact there. \\

    \textcolor{\sectioncolor}{\textbf{
    Is there an erratum?
    }
    If so, please provide a link or other access point.
    } \\
    
    No \\

    \textcolor{\sectioncolor}{\textbf{
    Will the dataset be updated (e.g., to correct labeling errors, add new
    instances, delete instances)?
    }
    If so, please describe how often, by whom, and how updates will be
    communicated to users (e.g., mailing list, GitHub)?
    } \\
    
    The dataset may be updated if any problematic instances are discovered or the label errors are corrected. In this case, we will issue a new version of the dataset to ensure past results are contextualized under the version of the dataset that was used. \\

    \textcolor{\sectioncolor}{\textbf{
    If the dataset relates to people, are there applicable limits on the
    retention of the data associated with the instances (e.g., were individuals
    in question told that their data would be retained for a fixed period of
    time and then deleted)?
    }
    If so, please describe these limits and explain how they will be enforced.
    } \\
    
    There is no limit. \\

    \textcolor{\sectioncolor}{\textbf{
    Will older versions of the dataset continue to be
    supported/hosted/maintained?
    }
    If so, please describe how. If not, please describe how its obsolescence
    will be communicated to users.
    } \\
    
    Yes, we will continue to host older versions of the dataset.
    This is an inherent feature of the Harvard Dataverse.
    \\

    \textcolor{\sectioncolor}{\textbf{
    If others want to extend/augment/build on/contribute to the dataset, is
    there a mechanism for them to do so?
    }
    If so, please provide a description. Will these contributions be
    validated/verified? If so, please describe how. If not, why not? Is there a
    process for communicating/distributing these contributions to other users?
    If so, please provide a description.
    } \\
    
    Yes, the dataset is licensed permissively, and the code used to create Wake Vision is open source. \\

    \textcolor{\sectioncolor}{\textbf{
    Any other comments?
    }} \\
    
    No additional comments \\